\pdfoutput=1
\documentclass[11pt]{article}

\usepackage[final]{acl}
\usepackage{arydshln}
\usepackage{times}
\usepackage{latexsym}

\usepackage[T1]{fontenc}

\usepackage[utf8]{inputenc}

\usepackage{microtype}

\usepackage{inconsolata}

\usepackage{tikz}
\usepackage{graphicx}
\usepackage{multirow}
\usepackage{amsmath}
\usepackage{amssymb}
\usepackage{wrapfig}   
\usepackage{float}
\usepackage{algorithm}
\usepackage{algorithmic}
\usepackage{mdframed}
\usepackage{tcolorbox}
\usepackage{subcaption}
\usepackage{listings}
\usepackage{lipsum}
\usepackage{array}
\usepackage{chngcntr}
\usepackage{ragged2e}
\tcbuselibrary{breakable}
\usepackage{booktabs}
\usepackage{makecell}
\usepackage{caption}
\usepackage{xcolor} 
\usepackage{pifont}  
\usepackage{longtable}

\usepackage{todonotes}

\usepackage[table]{xcolor} 
\definecolor{lightgray}{gray}{0.9} 
\tcbuselibrary{skins}
\usepackage{diagbox}

\usepackage{xstring}

\author{
    Jingsheng Zheng$^{\dag\ddag}$, Jintian Zhang$^{\dag\ddag}$, Yujie Luo$^{\dag\ddag}$, Yuren Mao$^{\dag}$, Yunjun Gao$^{\dag}$, \\
    \textbf{Lun Du}$^{\S\ddag}$, \textbf{Huajun Chen}$^{\dag\ddag}$, \textbf{Ningyu Zhang}$^{\dag\ddag}$\thanks{Corresponding Author.} \\
    $^{\dag}$Zhejiang University \quad $^{\S}$Ant Group \\
    $^{\ddag}$Zhejiang University - Ant Group Joint Laboratory of Knowledge Graph \\
    \texttt{zhengjohnson0@gmail.com, zhangningyu@zju.edu.cn}
}

\definecolor{myGreen}{RGB}{0, 150, 0}
\definecolor{myRed}{RGB}{200, 0, 0}

\definecolor{myblue}{RGB}{30, 144, 255} 

\usepackage{bm}
\newcommand{\perf}[2]{%
    #1%
    \rlap{$
        \,_{\IfBeginWith{#2}{-}%
            {\color{myGreen}\text{\tiny{(#2)}}}%
            {\color{myRed}\text{\tiny{(#2)}}}%
        }
    $}%
}

\tcbuselibrary{most} 

\newtcolorbox{definitionbox}[1][]{%
  enhanced,                
  title={Definition},      
  colback=blue!5,          
  colframe=blue!50!black,  
  coltitle=white,          
  colbacktitle=blue!60!black, 
  boxrule=1.5pt,           
  arc=3mm,                 
  fonttitle=\bfseries,     
  
  attach boxed title to top left={yshift=-2mm,xshift=4mm}, 
  boxed title style={
    rounded corners,
    boxrule=0pt,           
  },
  
  breakable,               
  #1                       
}

\usepackage{utfsym}
\usepackage{stfloats}
\newtcolorbox{insightbox}{
    float*=!hb,            
    width=\textwidth,    
    colback=myblue!5!white, 
    fontupper=\small\linespread{1.05}\selectfont,
    colframe=black,         
    boxrule=0.5pt,          
    arc=6pt,                
    left=8pt, right=8pt, top=4pt, bottom=4pt,
    before skip=10pt,
    after skip=10pt
}

\usepackage[scale=0.85]{FiraMono}
\newtcolorbox{promptbox}[2][]{
  enhanced,
  title={#2},
  colframe=myblue!50!gray,
  colback=myblue!5!white,
  colbacktitle=myblue!50!gray,
  fonttitle=\bfseries\large,
  fontupper=\ttfamily\small\linespread{1.1}\selectfont, 
  boxrule=1pt,
  arc=2mm,
  breakable,                
  top=10pt, bottom=10pt,    
  oversize,                 
  #1
}

\tcbuselibrary{raster}
\usepackage{fontawesome5}

\newtcolorbox{CaseStudyFrame}[2]{
    enhanced, 
    title={\large \textbf{#1}},
    colback=white, colframe=gray!70!black, coltitle=white,
    fonttitle=\bfseries,
    width=\textwidth, 
    boxrule=1pt,
    arc=4pt,
    label={#2}
}

\newcommand{\ContextBar}[1]{
    \begin{tcolorbox}[
        enhanced, colback=gray!10, colframe=gray!30, 
        sharp corners, boxrule=0.5pt,
        left=4pt, right=4pt, top=3pt, bottom=3pt
    ]
        \small #1 
    \end{tcolorbox}
    \vspace{0.3cm}
}

\newtcolorbox{SolutionColumn}[4]{
    enhanced,
    equal height group={#4}, 
    title={\textbf{#3}},
    colframe=#1, colback=white, coltitle=white,
    subtitle style={colback=#2, colupper=black},
    drop shadow, 
    width=\linewidth,
    fonttitle=\small\bfseries,
    left=3pt, right=3pt,
    valign=top 
}

\newtcolorbox{ReportFrame}[2]{
    enhanced,
    title={\large \textbf{#1}}, 
    colback=white, 
    colframe=gray!70!black, 
    coltitle=white,
    fonttitle=\bfseries,
    width=\textwidth,
    boxrule=1pt,
    arc=4pt,
    fontupper=\ttfamily\small\linespread{1.1}\selectfont, 
    label={#2}
}

\newtcolorbox{AlignedDesc}[1]{
    enhanced,
    frame hidden,      
    colback=white,     
    equal height group={#1_desc}, 
    left=0pt, right=0pt, top=0pt, bottom=5pt, 
    nobeforeafter      
}

\newcommand{\HumanView}[3][]{
    \begin{tcolorbox}[
        enhanced, 
        code={\ifstrempty{#1}{}{\tcbset{equal height group={#1_view}}}},
        colback=gray!5,       
        colframe=gray!30,     
        title={\footnotesize \faUserTie\ \textbf{Human Intuition}}, 
        boxrule=1pt, 
        left=3pt, right=3pt, top=3pt,
        coltitle=black
    ]
        \small \textit{"#2"} 
        \par\smallskip
        \textbf{Verdict:} \textbf{#3}
    \end{tcolorbox}
}

\newcommand{\WMView}[4][]{
    \begin{tcolorbox}[
        enhanced, 
        code={\ifstrempty{#1}{}{\tcbset{equal height group={#1_view}}}},
        colback=myblue!5,     
        colframe=myblue!30,   
        title={\footnotesize \faRobot\ \textbf{World Model #2}}, 
        boxrule=1pt, 
        left=3pt, right=3pt, top=3pt,
        coltitle=black 
    ]
        \small \textit{"#3"} 
        \par\smallskip
        \textbf{Verdict:} \textbf{#4}
    \end{tcolorbox}
}

\newcommand{\OutcomeFooter}[1]{
    \vspace{0.5cm}
    \hrule
    \vspace{0.1cm}
    \small 
    \textbf{Outcome:} #1
}

\usepackage[scale=0.85]{FiraMono} 

\newtcblisting{caserawbox}[2][]{
  enhanced,
  title={#2},
  colback=white,
  colframe=gray!70!black,
  coltitle=white,
  fonttitle=\bfseries\large,
  boxrule=1pt,
  arc=2mm,
  breakable,
  listing only,
  listing options={
    basicstyle=\ttfamily\scriptsize,
    breaklines=true,
    columns=fullflexible,
    keepspaces=true,
    showstringspaces=false,
    extendedchars=false, 
    literate={ }{ {~} }1 {≈}{{$\approx$}}1, 
    inputencoding=utf8,
  },
  #1
}

\usepackage{booktabs}
\usepackage{graphicx}
\usepackage{xcolor}
\usepackage{colortbl}

\definecolor{domainbg}{gray}{0.95}

\usepackage{xspace}
\newcommand{\ours}{\textsc{ForeAgent}\xspace}

\title{Can We Predict Before Executing Machine Learning Agents?}

\begin{document}
\maketitle

\begin{abstract}
    Autonomous machine learning agents have revolutionized scientific discovery, yet they remain constrained by a Generate-Execute-Feedback paradigm. Previous approaches suffer from a severe Execution Bottleneck, as hypothesis evaluation relies strictly on expensive physical execution. To bypass these physical constraints, we internalize execution priors to substitute costly runtime checks with instantaneous predictive reasoning, drawing inspiration from World Models. In this work, we formalize the task of Data-centric Solution Preference and construct a comprehensive corpus of 18,438 pairwise comparisons. We demonstrate that LLMs exhibit significant predictive capabilities when primed with a Verified Data Analysis Report, achieving 61.5\% accuracy and robust confidence calibration. Finally, we instantiate this framework in \ours, an agent that employs a Predict-then-Verify loop, achieving a 6x acceleration in convergence while surpassing execution-based baselines by +6\%. Our code and dataset are publicly available at \url{https://github.com/zjunlp/predict-before-execute}.
\end{abstract}

\section{Introduction}

Autonomous machine learning agents have emerged as powerful tools for solving complex challenges in scientific discovery~\cite{Wen-lin2025, chen2025largelanguagemodelbaseddata}.
Mainstream frameworks~\cite{jiang2025aideaidrivenexplorationspace, ou2025automindadaptiveknowledgeableagent} typically rely on an iterative \textit{``Generate-Execute-Feedback''} loop where the system refines code based on runtime output~\cite{yao2023reactsynergizingreasoningacting}.
However, this paradigm suffers from a severe \textbf{Execution Bottleneck} as physical execution is computationally expensive and slow, often consuming up to 9 hours per run in benchmarks like MLE-Bench~\cite{chan2025mlebenchevaluatingmachinelearning}.
Increasingly, recent research has identified this latency issue and sought to mitigate the computational overhead through heuristic pruning strategies~\cite{trirat2025automlagentmultiagentllmframework, kulibaba2025kompeteaiacceleratedautonomousmultiagent}.

\begin{figure}[t]
    \centering
    \includegraphics[width=1\linewidth]{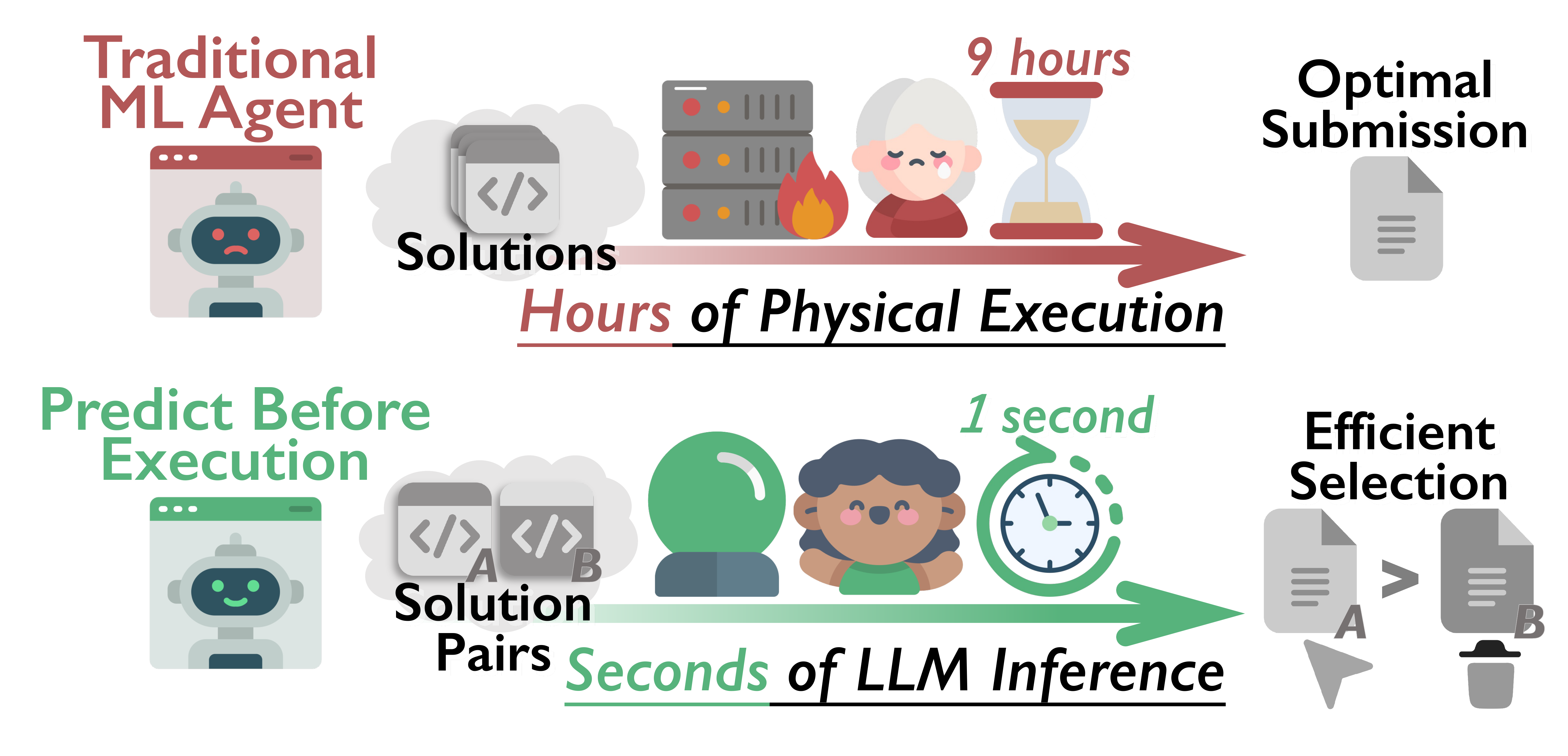}
    \caption{\textbf{From Execution to Inference.}
    Traditional ML agents improve through costly execution and external feedback, incurring substantial latency. 
    Our work investigates whether superior data-grounded solutions can be identified before execution by leveraging ``Implicit Execution Priors''.}
    \label{fig:top_concept}
\end{figure}

To fundamentally bypass these physical constraints, the concept of \textbf{World Models}~\cite{Ding2025} offers a transformative alternative (Figure~\ref{fig:top_concept}).
Originating from reinforcement learning, world models enable agents to simulate environmental dynamics and evaluate actions via internal predictions rather than external trials~\cite{ha2018worldmodels, hafner2024masteringdiversedomainsworld}.
Recent advancements have extended this capability to the code domain by predicting execution outputs directly~\cite{li2025codeiocondensingreasoningpatterns, faircodegenteam2025cwmopenweightsllmresearch}.
Motivated by this, we explore whether agents can internalize execution priors, substituting costly runtime checks with instantaneous predictive reasoning.
The potential to replace \textbf{9 hours} of physical latency with \textbf{1 second} of neural speed brings us to a fundamental question: 
\textbf{\textit{Can we compress hours of physical execution into seconds of logical inference?}}

To answer this question, we formalize the task of \textbf{Data-centric Solution Preference}, where the model must predict the relative performance of two algorithmic solutions given a data analysis report, through reasoning without physical execution.
To rigorously evaluate this, we construct a large-scale corpus comprising 18,438 pairwise comparisons.
Our main experiments yield strong evidence: \textbf{LLMs exhibit significant predictive capabilities}, with DeepSeek-V3.2-Thinking achieving 61.5\% accuracy, outperforming both random guessing (50.0\%) and complexity-based heuristics (50.8\%).
Further analysis reveals that reasoning-optimized architectures transcend complexity heuristics through genuine data reasoning, yielding well-calibrated confidence that ensures the reliability of implicit evaluation.
Finally, we integrate this predictive mechanism into \textbf{\ours}, an agent that employs a \textit{Predict-then-Verify} loop to decouple exploration from execution, expanding the search space by $3.2\times$ and achieving a $6\times$ acceleration while delivering a +6\% performance gain over standard baselines.

In summary, our contributions are three-fold:
\begin{itemize}
    \item We define the novel task of \textbf{Data-centric Solution Preference} and construct a comprehensive corpus of 18,438 pairs, answering the titular question that \textbf{LLMs Exhibit Significant Predictive Capabilities.}
    \item We operationalize this framework in \textbf{\ours}, an agent that employs a \textit{Predict-then-Verify} loop to decouple exploration from execution, enabling it to expand the search space by $\bm{3.2\times}$ and achieve a $\bm{6\times}$ acceleration and a \textbf{+6\%} performance gain over the baseline.
    \item We contribute a large-scale \textbf{Open-Source Dataset} of verified execution trajectories, serving as a foundational corpus for training scalable Reward Models to accelerate reinforcement learning rollouts and optimization across diverse agent frameworks.
\end{itemize}

\begin{figure*}[t]
    \centering
    \includegraphics[width=\linewidth]{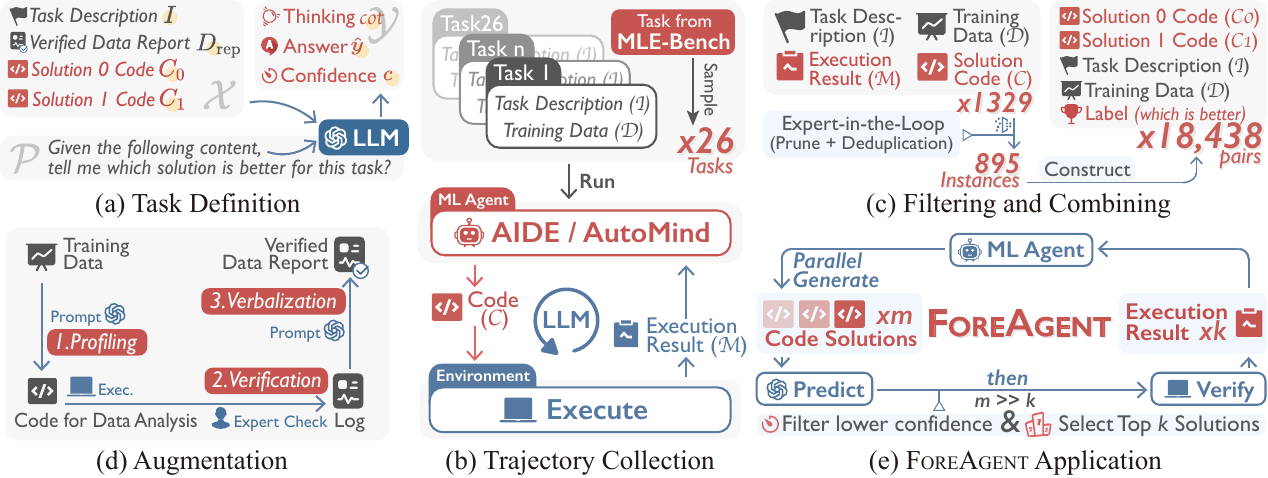}
    \caption{\textbf{Overview of the Framework.} \textbf{(a) Task Definition:} The \textit{Data-centric Solution Preference} task predicts solution superiority and confidence via latent reasoning. \textbf{(b-c) Data Curation:} We collect and filter real-world agent trajectories to construct the \textit{Preference Corpus}. \textbf{(d) Augmentation:} Inputs are augmented with \textit{Verified Data Reports} via a ``Profile-Verify-Verbalize'' pipeline. \textbf{(e) \ours Application:} The model serves as a filter within the \textit{Predict-then-Verify} loop, predicting preference \textit{before} physical execution to prune candidates.}
    \label{fig:main}
\end{figure*}

\section{Background}
\label{sec:background}

\subsection{The Paradigm of Autonomous ML Agents}
\label{sec:agent_paradigm}

An autonomous Machine Learning (ML) task aims to generate an optimal solution code $C^*$ from the code space $C$ that maximizes a metric $M$ on a dataset $\mathcal{D}$, given a natural language instruction $I$ (see Appendix Figure~\ref{fig:task_instruction_denoising}):
\begin{equation}
    C^* = \operatorname*{arg\,max}_{C} M(I, C, \mathcal{D})
\end{equation}
Current agents typically follow a \textit{Generate-Execute-Feedback} paradigm~\cite{zhu2025surveydataagentsemerging}.
For instance, \textbf{AIDE}~\cite{jiang2025aideaidrivenexplorationspace} organizes solution exploration as a tree search process involving sequential drafting, debugging, and iterative improvement via execution feedback.
Building upon this, \textbf{AutoMind}~\cite{ou2025automindadaptiveknowledgeableagent} integrates a curated expert knowledge base with a self-adaptive coding strategy to tackle more intricate problems (see Appendix~\ref{app:extended_related_work} for details).

\begin{table}[t]
    \centering
    \footnotesize
    \setlength{\tabcolsep}{3pt} 
    \renewcommand{\arraystretch}{1.2}
    \begin{tabular}{l p{3cm} r r r}
        \toprule
        \textbf{Domain} & \textbf{Paradigms} & \textbf{\# Tsk} & \textbf{\# Sols} & \textbf{\# Pairs} \\
        \midrule
        \textbf{CV} & Classification, Segmentation, Generation, Restoration & 9 & 289 & 5,952 \\
        \textbf{NLP} & Classification, Matching, QA, Sequence Labeling, Ranking & 8 & 303 & 6,682 \\
        \multirow{2}{*}{\textbf{\makecell[l]{Data\\Science}}} & Regression, Time-Series, Audio, Tabular, Grading & 9 & 303 & 5,804 \\
        \midrule
        \textbf{Total} & \textit{26 Distinct Tasks across 3 Domains} & \textbf{26} & \textbf{895} & \textbf{18,438} \\
        \bottomrule
    \end{tabular}
    \caption{Statistics of the Preference Corpus. We aggregate 26 tasks into three primary domains, ensuring a balanced distribution of $\sim$6,000 pairs each. (See Appendix~\ref{sec:appendix_task_detail} for granular breakdown).}
    \label{tab:task_distribution}
\end{table}



\subsection{The Execution Bottleneck}
\label{sec:motivation}

The primary constraint in current agents is the reliance on physical execution for feedback.
Formally, the update of solution $C_{t+1}$ depends on the result $R_t$ from executing on dataset $\mathcal{D}$:
\begin{equation}
    C_{t+1} \leftarrow \text{Agent}(I, C_t, \underbrace{\text{Execute}(C_t, \mathcal{D})}_{R_t})
\end{equation}
Unlike symbolic tasks with instantaneous verification, training deep learning models involves heavy computation, frequently leading to timeout failures~\cite{chan2025mlebenchevaluatingmachinelearning}.
This efficiency gap \textbf{necessitates compressing hours of physical execution into seconds of logical inference}, mirroring how human experts utilize mental simulation to discard sub-optimal algorithms prior to implementation.

\subsection{Implicit World Modeling in Data Domains}
\label{sec:implicit_modeling}

We investigate whether LLMs can function as an \textit{Implicit World Model}~\cite{ha2018worldmodels, hafner2024masteringdiversedomainsworld}.
While recent works explore this direction across diverse symbolic and interactive domains~\cite{li2025wordworldlargelanguage, faircodegenteam2025cwmopenweightsllmresearch, Just2024datacentrichuman}, our \textbf{Data-centric Solution Preference} task is distinct:
unlike tracking explicit states, the model must anticipate the invisible coupling of algorithmic logic and stochastic data.
Thus, we formulate the problem as a \textit{Pairwise Preference} task~\cite{Shen2024towardsdatacentricrlhf}, determining the superior solution purely via reasoning to identify promising candidates prior to execution.

\section{Preference Corpus Curation}
\label{sec:data-construction}

This section details the curation of our preference corpus. We begin by formalizing the task to clarify the data requirements, followed by the collection and augmentation processes.

\subsection{Task Definition}
\label{sec:task_definition}

We model the data-centric task as a pairwise selection task: given a task description, a data report, and two candidate solutions, the objective is to identify the superior solution and estimate a confidence score (Figure~\ref{fig:main}(a)).
Formally, the input $\mathcal{X}$ is:
\begin{equation}
    \mathcal{X} = \left( I, D_{rep}, \{C_0, C_1\}, \mathcal{P} \right)
\end{equation}
where $I$, $D_{rep}$, $\{C_0, C_1\}$, and $\mathcal{P}$ denote the task, data report, code pair, and system prompt, respectively. The output $\mathcal{Y}$ is defined as:
\begin{equation}
    \mathcal{Y} = \left\{ (cot, \hat{y}, c) \mid cot, \; \hat{y} \in \{0, 1\}, \; c \in [0, 1.0] \right\}
\end{equation}
consisting of the reasoning $cot$, predicted winner $\hat{y}$, and confidence $c$, which serves as the gating threshold in Section~\ref{sec:agent}.

\subsection{Source and Scope}
\label{sec:source_scope}

To instantiate the task inputs defined above, we construct a large-scale corpus derived from the real-world execution trajectories of two ML agents, \textbf{AIDE}~\cite{jiang2025aideaidrivenexplorationspace} and \textbf{AutoMind}~\cite{ou2025automindadaptiveknowledgeableagent}, operating on \textbf{MLE-bench}~\cite{chan2025mlebenchevaluatingmachinelearning} platform (Figure~\ref{fig:main}(b)).
Powered by DeepSeek-V3.1~\cite{deepseekai2025deepseekv3technicalreport} and o3-mini~\cite{openai2025o3card}, these agents generate \textbf{1,329} valid solutions across \textbf{26} diverse tasks (Table~\ref{tab:task_distribution}).
Unlike synthetic snippets, these candidates represent \textit{complete ML workflows} that are entirely generated by agents and absent from any pre-training corpora, heavily incorporating logically incomplete but executable intermediate states to model noisy real-world exploration (see Appendix \ref{sec:trajectory_sampling}).
Therefore, identifying the superior solution requires evaluating \textit{how well an algorithm fits the specific data characteristics}, rather than merely checking for code syntax.

\subsection{Dataset Curation and Instantiation}
\label{sec:data_construction}

For rigorous evaluation, we implement an \textbf{Expert-in-the-Loop} pipeline to prune raw trajectories into \textbf{895} high-quality instances. 
This process involves deduplication, taxonomy tagging, and expert sampling to cap dominant methods and ensure algorithmic diversity. Next, we instantiate the dataset by exhaustively generating pairwise combinations from this curated corpus. 
We apply strict filtering to discard ambiguous pairs and balance the ground-truth winner's position to mitigate position bias~\cite{Shi2024judingthejudges}. 
This yields a final dataset of \textbf{18,438 comparisons} (Figure~\ref{fig:main}(c)), utilizing \textbf{micro-averaged accuracy} as the primary metric.

\subsection{Input Augmentation: The Verified Data Analysis Report}
\label{sec:input_augmentation}

To address LLMs' numerical limitations~\cite{davies2025languagemodelsembednumbers, li-etal-2025-exposing} and context constraints preventing direct data ingestion, we augment inputs with a \textbf{Verified Data Analysis Report} that transforms raw statistics into semantic narratives~\cite{Rytting2021leveragingtheinductivebias, zhang2025srllmrethinkingstructuredrepresentation}.
To guarantee factual grounding and prevent hallucination, we implement a strict protocol (Figure~\ref{fig:main}(d)) consisting of three concrete steps:
(1) \textbf{Code Generation}: GPT-5.1~\cite{openai2025gpt5card} generates a Python script to profile raw data with labels masked.
For example, it writes ``\texttt{print(df['target'].value\_counts())}'' to inspect the target distribution.
(2) \textbf{Execution and Verification}: The script runs in a sandbox to produce standard output.
A human expert performs a strict pass or fail validity check to ensure the log is free of runtime errors.
For instance, the execution log returns a raw fact like ``\texttt{Target Distribution: 0: 0.915, 1: 0.085}''.
(3) \textbf{Verbalization}: GPT-5.1 reads this execution log and translates it into a semantic insight.
Following the previous example, it produces the final report: ``Data Imbalance Warning: Severe class imbalance (Pos: 8.5\%). Implication: Accuracy is not a suitable metric; consider using F1-score.''
This process ensures reliable semantic grounding for the task (see case in Appendix Figure~\ref{fig:report_patent}).

\section{Main Experiments}
\label{sec:experiments}

\begin{table*}[!ht]
    \centering
    \footnotesize 
    \renewcommand{\arraystretch}{1.2} 
    \setlength{\tabcolsep}{1.1pt} 
    
    \newcommand{\sres}[2]{{\scriptsize $#1_{\pm #2}$}}
    \newcommand{\res}[2]{$#1_{\pm #2}$}
    \newcommand{\bsres}[2]{{\scriptsize $\bm{#1}_{\pm \bm{#2}}$}}
    \newcommand{\bres}[2]{$\bm{#1}_{\pm \bm{#2}}$}
    
    \begin{tabular}{ll | ccc | ccc | ccc | c}
        \toprule
        \multicolumn{2}{r|}{\hfill \textit{\textbf{(Acc. \%)}} \textbf{Task Dims.} $\rightarrow$\hspace{0.5em}} & \multicolumn{3}{c|}{\textit{Domain}} & \multicolumn{3}{c|}{\textit{Difficulty}} & \multicolumn{3}{c|}{\textit{Task Paradigm}} & \textbf{Sols.} \\
        
        \cmidrule(lr){3-5} \cmidrule(lr){6-8} \cmidrule(lr){9-11}
        
        \multicolumn{2}{l|}{\hspace{0.2em}$\downarrow$ \textbf{Sols. Attrs.} \hfill } & CV & NLP & Data Sci. & Easy & Med. & Hard & Class. & Regres. & Others & \textbf{Avg Acc} \\ 
        \midrule

        \rowcolor{myblue!8} 
        \multicolumn{12}{c}{\textbf{\raisebox{-0.15em}{\includegraphics[height=0.9em]{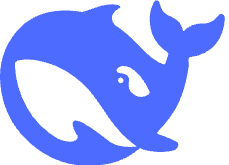}} DeepSeek-V3.2 (Thinking mode)}} \\ 
        \midrule
        
        \rowcolor{myblue!4}
          & Traditional & \sres{60.2}{0.9} & \sres{70.6}{0.6} & \sres{59.3}{0.5} & \sres{59.8}{1.1} & \sres{69.1}{0.2} & \sres{61.1}{0.7} & \sres{61.5}{0.5} & \sres{61.2}{0.7} & \sres{76.2}{0.5} & \cellcolor{myblue!10}\bres{64.5}{0.6} \\
        
        \rowcolor{myblue!4}
        \multirow{-2}{*}{\textit{Algo Era}} 
          & Modern      & \sres{59.1}{0.5} & \sres{65.0}{0.1} & \sres{56.3}{0.5} & \sres{60.7}{0.3} & \sres{61.7}{0.3} & \sres{55.1}{0.6} & \sres{57.8}{0.2} & \sres{62.5}{0.4} & \sres{62.3}{0.5} & \cellcolor{myblue!10}\res{60.4}{0.2} \\
        \cmidrule(l){1-12}
        
        \rowcolor{myblue!4}
          & Cross-Algo  & \sres{56.6}{0.3} & \sres{68.9}{0.7} & \sres{58.4}{0.9} & \sres{57.6}{1.0} & \sres{68.2}{0.4} & \sres{57.7}{1.5} & \sres{59.8}{0.7} & \sres{60.6}{0.7} & \sres{74.1}{0.9} & \cellcolor{myblue!10}\bres{62.8}{0.6} \\
        
        \rowcolor{myblue!4}
        \multirow{-2}{*}{\textit{Granularity}} 
          & Self-Comp.  & \sres{60.1}{0.6} & \sres{65.1}{0.2} & \sres{56.3}{0.8} & \sres{61.6}{0.3} & \sres{60.9}{0.4} & \sres{56.5}{1.0} & \sres{58.2}{0.1} & \sres{62.9}{0.4} & \sres{62.1}{0.5} & \cellcolor{myblue!10}\res{60.7}{0.1} \\
        \cmidrule(l){1-12}
        
        \rowcolor{myblue!4}
          & Low         & \sres{57.6}{0.4} & \sres{69.8}{0.5} & \sres{57.2}{0.3} & \sres{58.9}{0.6} & \sres{66.2}{0.2} & \sres{58.9}{0.7} & \sres{58.6}{0.2} & \sres{61.6}{0.2} & \sres{73.3}{0.9} & \cellcolor{myblue!10}\bres{62.1}{0.3} \\
        
        \rowcolor{myblue!4}
          & Medium      & \sres{59.6}{0.3} & \sres{65.1}{0.1} & \sres{58.1}{0.2} & \sres{60.5}{0.2} & \sres{63.3}{0.1} & \sres{56.6}{0.2} & \sres{58.1}{0.2} & \sres{63.4}{0.3} & \sres{64.6}{0.6} & \cellcolor{myblue!10}\res{61.3}{0.1} \\
          
        \rowcolor{myblue!4}
        \multirow{-3}{*}{\textit{Complexity}} 
          & High        & \sres{61.2}{2.0} & \sres{80.1}{0.7} & \sres{50.0}{1.1} & \sres{76.8}{2.8} & \sres{58.4}{1.6} & \sres{52.7}{0.9} & \sres{60.3}{2.5} & \sres{58.4}{1.4} & \sres{61.3}{1.7} & \cellcolor{myblue!10}\res{59.6}{1.4} \\
        
        \midrule 
        \rowcolor{myblue!10}
        \multicolumn{2}{l|}{\textbf{Tasks Avg Acc}} 
        & \res{59.3}{0.5} & \bres{66.9}{0.2} & \res{57.4}{0.2} & \res{60.4}{0.5} & \bres{63.9}{0.2} & \res{57.0}{0.3} & \res{58.9}{0.3} & \res{62.1}{0.1} & \bres{66.8}{0.5} & \cellcolor{myblue!20}\textbf{\bres{61.5}{0.2}} \\ 

        \midrule
        \addlinespace[0.5em]

        \multicolumn{12}{c}{\textbf{\raisebox{-0.1em}{\includegraphics[height=0.9em]{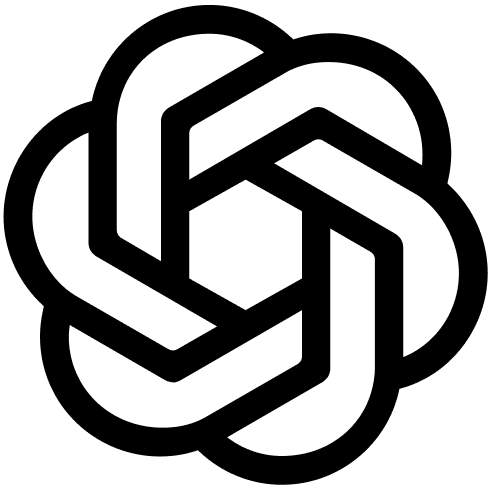}} GPT-5.1}} \\ 
        \midrule
        
        \multirow{2}{*}{\textit{Algo Era}} 
          & Traditional & \sres{60.1}{0.4} & \sres{64.7}{0.2} & \sres{59.5}{0.2} & \sres{56.6}{0.4} & \sres{65.5}{0.2} & \sres{63.4}{0.3} & \sres{59.3}{0.6} & \sres{62.2}{0.2} & \sres{67.2}{0.3} & \cellcolor{gray!10}\bres{62.0}{0.2} \\
          & Modern      & \sres{54.5}{0.2} & \sres{62.2}{0.8} & \sres{56.3}{0.1} & \sres{55.1}{0.4} & \sres{59.9}{0.4} & \sres{57.9}{0.0} & \sres{56.4}{0.5} & \sres{58.2}{0.5} & \sres{59.8}{0.6} & \cellcolor{gray!10}\res{57.7}{0.3} \\
        \cmidrule(l){1-12}
        
        \multirow{2}{*}{\textit{Granularity}} 
          & Cross-Algo  & \sres{58.0}{1.1} & \sres{61.2}{0.3} & \sres{56.7}{0.1} & \sres{55.3}{0.7} & \sres{61.3}{0.2} & \sres{59.8}{0.1} & \sres{55.7}{0.8} & \sres{61.7}{0.3} & \sres{62.0}{0.5} & \cellcolor{gray!10}\bres{59.0}{0.3} \\
          & Self-Comp.  & \sres{54.8}{0.0} & \sres{64.2}{1.0} & \sres{57.7}{0.1} & \sres{55.3}{0.4} & \sres{61.6}{0.4} & \sres{59.2}{0.2} & \sres{58.0}{0.7} & \sres{57.6}{0.5} & \sres{62.2}{0.7} & \cellcolor{gray!10}\res{58.7}{0.3} \\
        \cmidrule(l){1-12}
        
        \multirow{3}{*}{\textit{Complexity}} 
          & Low         & \sres{56.4}{0.2} & \sres{66.6}{0.5} & \sres{56.2}{0.2} & \sres{56.8}{0.5} & \sres{64.2}{0.2} & \sres{57.6}{0.2} & \sres{57.9}{0.7} & \sres{59.3}{0.2} & \sres{68.7}{0.3} & \cellcolor{gray!10}\bres{60.1}{0.3} \\
          & Medium      & \sres{55.8}{0.2} & \sres{60.6}{0.6} & \sres{59.3}{0.2} & \sres{54.6}{0.4} & \sres{61.2}{0.3} & \sres{61.1}{0.2} & \sres{56.5}{0.5} & \sres{60.0}{0.5} & \sres{60.5}{0.6} & \cellcolor{gray!10}\res{58.6}{0.3} \\
          & High        & \sres{50.8}{0.3} & \sres{79.0}{0.8} & \sres{57.2}{1.5} & \sres{44.2}{2.7} & \sres{56.2}{0.4} & \sres{59.7}{1.6} & \sres{56.0}{0.7} & \sres{54.5}{0.7} & \sres{56.2}{1.1} & \cellcolor{gray!10}\res{55.3}{0.3} \\
        
        \midrule
        \rowcolor{gray!10}
        \multicolumn{2}{l|}{\textbf{Tasks Avg Acc}} 
        & \res{55.4}{0.2} & \bres{63.0}{0.6} & \res{57.4}{0.1} & \res{55.5}{0.4} & \bres{61.6}{0.3} & \res{59.7}{0.1} & \res{57.2}{0.5} & \res{59.2}{0.3} & \bres{62.2}{0.5} & \cellcolor{gray!20}\textbf{\res{58.8}{0.3}} \\
        
        \bottomrule
    \end{tabular}

    \caption{\textbf{Main Results: Predictive Capability and Boundary Analysis.} 
    This table presents the Pairwise Preference Accuracy (\%) of the evaluated LLMs averaged over three runs, stratified by Task Dimensions and Solution Attributes. Results are reported as Mean $_{\pm \text{Stdev}}$. 
    \textbf{DeepSeek-V3.2 (Thinking Mode)} and \textbf{GPT-5.1} achieve global averages of \textbf{61.5\%} and \textbf{58.8\%} respectively, significantly outperforming the random baseline of \textbf{50\%} and the complexity-based heuristic baseline of \textbf{50.8\%}.} 
    
    \label{tab:final_compact_matrix_clean}
    
\end{table*}

\subsection{Experimental Setup}
\label{sec:setup}

\paragraph{Models and Inference Configuration.}
We evaluate two state-of-the-art models: \textbf{DeepSeek-V3.2-Thinking}~\cite{deepseekai2025deepseekv3technicalreport} and \textbf{GPT-5.1} (gpt-5.1-2025-11-13)~\cite{openai2025gpt5card} with reasoning instructions~\cite{wei2023chainofthoughtpromptingelicitsreasoning, kojima2023largelanguagemodelszeroshot}, adhering to the task in Section~\ref{sec:task_definition}.
Following provider guidelines~\cite{deepseek2024apidocs}, we set the temperature $\tau = 1.0$ for both models as the recommended default for data analysis.

\paragraph{Metrics and Baselines.}
The primary metric is \textbf{Micro-Averaged Accuracy} across 18,438 pairwise comparisons.
We benchmark against two baselines:
(1) \textbf{Random Guess (50.0\%)};
(2) \textbf{Complexity Heuristic (50.8\%)}: A rule-based baseline that assumes ``complex is better''.
To operationalize this, we employed an LLM to score each solution (1-10) across three dimensions: \textit{Code Engineering}, \textit{Model Architecture}, and \textit{Data Pipeline} (see Appendix Figure~\ref{fig:appendix_complexity_prompt}).
This baseline predicts the winner based on the aggregate complexity score.

\subsection{Main Results: Feasibility of Run-Free Preference}
\label{sec:main_results}

The stratified pairwise accuracy results in Table~\ref{tab:final_compact_matrix_clean} validate the feasibility of our approach.

\textbf{LLMs Exhibit Significant Predictive Capabilities.}
Both models significantly outperform the random baseline and the complexity heuristic with statistical significance, with \textit{DeepSeek-V3.2-Thinking} achieving \textbf{61.5\%} and \textit{GPT-5.1} achieving \textbf{58.8\%}.
This performance gap ($>10\%$) proves LLMs derive valid signals from static inputs through genuine reasoning rather than heuristics, despite the task remaining a challenging frontier.


\section{Analysis \& Insights}
\label{sec:analysis}

\begin{figure*}[!ht]
    \centering
    \includegraphics[width=1\linewidth]{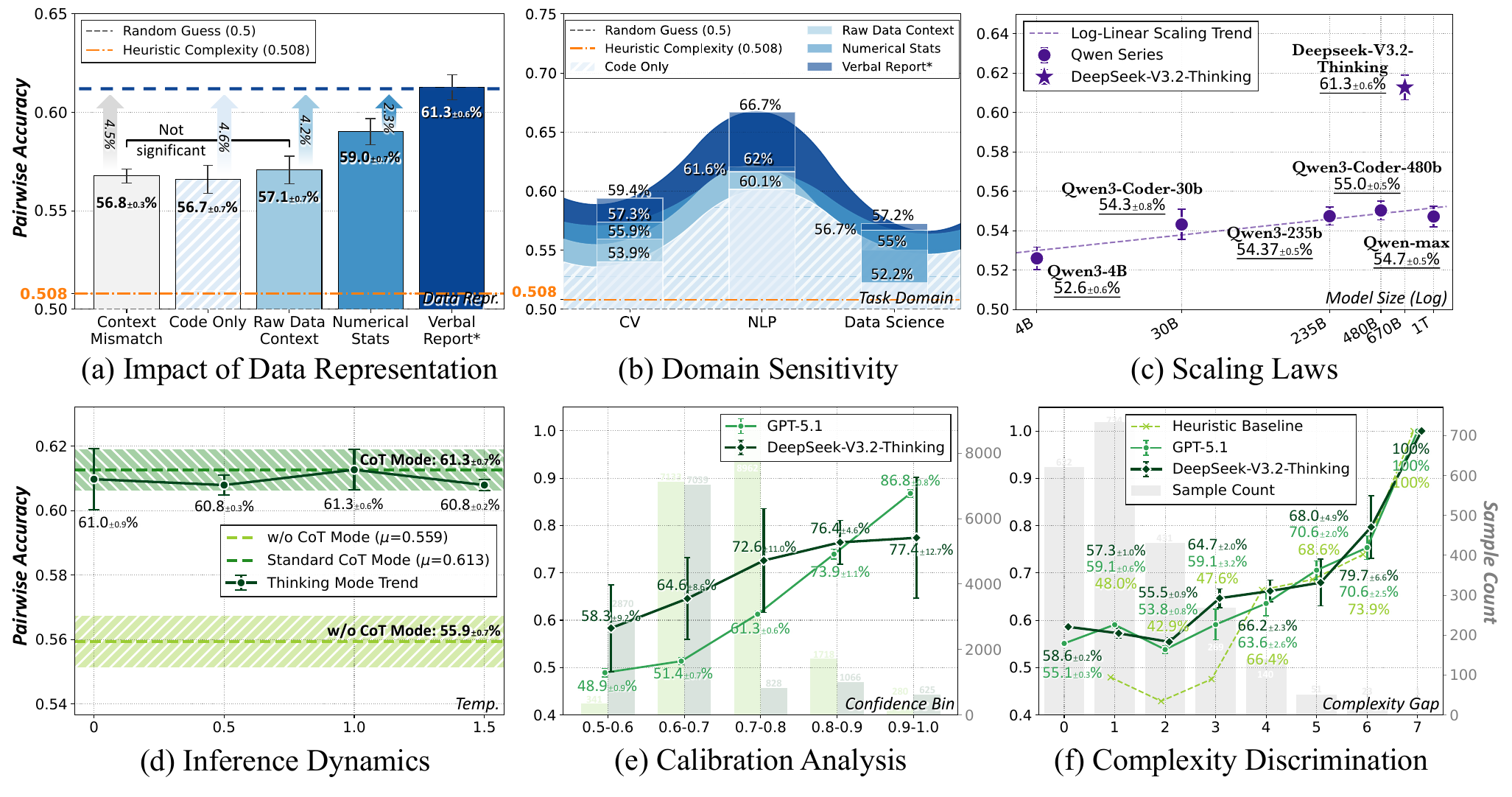}
    \caption{\textbf{Comprehensive Analysis of World Model Mechanisms and Capabilities.}
\textbf{(a) Impact of Data Representation:} Predictive success stems from semantic data understanding rather than complexity heuristics.
\textbf{(b) Domain Sensitivity:} The superiority of verbal reports remains consistent across domains.
\textbf{(c) Scaling Laws:} Accuracy decouples from pure parameter scaling.
\textbf{(d) Inference Dynamics:} Active reasoning outperforms direct answering with robust stability across temperatures.
\textbf{(e) Calibration Analysis:} Self-reported confidence correlates with accuracy.
\textbf{(f) Complexity Discrimination:} Accuracy scales with the complexity gap.}
    \label{fig:analysis}
\end{figure*}

\begin{insightbox}
    \textbf{{\usym{2660}} Discussion~\ref{subsec:rq1_representation}. The Gap Between Semantic and Numeric Spaces.}
    We observe that raw data yields insignificant gains over code-only input.
    We attribute this to the fact that continuous numerical values are \textbf{Weak-Semantic Symbols} that lack generalizable topological structure in the embedding space~\cite{davies2025languagemodelsembednumbers}.
    While language serves as a \textbf{Strong-Structural Symbol} carrying the compressed, empirical representations of human reasoning~\cite{Rytting2021leveragingtheinductivebias}, raw numbers appear to the model as unstructured high-entropy noise.
    Our verbalization strategy bridges this gap by projecting numeric data into the semantic manifold, injecting necessary inductive bias.
    Looking ahead, a fundamental resolution requires integrating \textbf{Symbolic Regression}~\cite{grayeli2024symbolicregressionlearnedconcept} to distill intricate logic directly from data.
\end{insightbox}

In this section, we deconstruct the mechanisms of the ``Implicit World Model'' through four pivotal research questions to answer: \textit{why can reasoning substitute for execution, and to what extent?}

While our representational analysis utilizes the full dataset, subsequent analysis (RQ2--RQ3) employs a focused subset capped at 15 solutions per task (2,292 pairs).
For ranking evaluations, we sample 105 instances per task to align with the pairwise baseline complexity ($C(15,2)$).

\subsection{RQ1: The Cognitive Mechanism of Data Representation}
\label{subsec:rq1_representation}

To distinguish genuine causal reasoning from syntactic memorization, we conducted a systematic study on input modalities (Figure~\ref{fig:analysis}(a)).
We instantiated four progressively enriched levels: \textit{Code Only} (task description + code), \textit{Raw Data} (appending initial samples), \textit{Numerical Stats} (execution logs from data analysis scripts), and \textit{Verbal Report} (full semantic analysis), alongside a \textit{Context Mismatch} control (pairing code with irrelevant context).

\paragraph{Finding 1: Predictive Success Stems from Semantic Data Understanding, Not Simple Complexity Heuristics.}
Our results refute a potential concern that LLMs merely rely on a complexity heuristic.
Figure~\ref{fig:analysis}(a) demonstrates a clear performance progression from the Heuristic Baseline (50.8\%) and Code Only (56.7\%) to Numerical Stats (59.0\%), peaking with Verbal Reports (61.3\%).
The insignificant gain of the Context Mismatch (56.8\%) over Code Only confirms that predictive success hinges on strict \textit{Semantic Alignment}.
The superiority of verbal narratives over raw statistics reveals that models operate primarily as \textit{rhetorical reasoners}, triggering an inference jump that is consistent across domains (Figure~\ref{fig:analysis}(b)).








\begin{insightbox}
    \textbf{{\textcolor{red}{\usym{2665}}} Discussion~\ref{subsec:rq3_scaling}: The Illusion of Scaling on World-Blind Models.}
    We attribute scaling failures to \textbf{Information Scarcity}: static training corpora consist of code paired with merely \textit{trivial inputs} or abstract descriptions, lacking the \textit{large-scale data distributions} required for true dynamic execution interplay~\cite{liu2022mindseyegroundedlanguage}.
    While our results confirm that models can exploit sparse dynamic traces hidden in existing data, they remain fundamentally \textbf{World-Blind Learners}~\cite{floridi2025categoricalanalysislargelanguage}.
    To transcend this mere ``syntactic mimicry'', future scaling must pivot from static ingestion to \textbf{Interactive Simulation}, grounding agents in genuine causal feedback loops~\cite{benderkoller2020climbingtowardsnlu}.
\end{insightbox}

\subsection{RQ2: Capabilities, Boundaries, and Algorithmic Bias}
\label{subsec:rq2_boundary}

In this section, we  analyze the necessity of reasoning, domain sensitivity, generalization to ranking, and the reliability of its confidence.

\paragraph{Finding 2: Reasoning Unlocks Capabilities, Yet Distinct Cognitive Boundaries Persist Across Domains.}

Figure~\ref{fig:analysis}(d) identifies \textit{reasoning} as the primary engine, with the Thinking Mode (CoT)~\cite{deepseekai2025deepseekr1incentivizingreasoningcapability, openai2024openaio1card} ($61.3\%$) outperforming Direct Answering ($55.9\%$).
This performance remains robust across temperatures ($T \in [0, 1.5]$) and trajectory variance (see Appendix \ref{app:trajectory_variance}), implying an invariant logical core that relies on genuine reasoning rather than exploiting easy artifacts from the same agent run.
However, this capability is constrained by the problem landscape; Table~\ref{tab:final_compact_matrix_clean} reveals sharp performance stratifications across the Task-Solution matrix.
On the \textit{Task Dimension}, models demonstrate a preference for NLP ($66.9\%$) and Easy ($63.9\%$) paradigms.
Simultaneously, analysis on the \textit{Solution Dimension} reveals a comparison preference for Traditional ML within the \textit{Algo Era} ($64.5\%$), a ``Complexity Tax'' ($59.6\%$ on complex code), and a \textit{Granularity} bottleneck, where the model is more effective at distinguishing broad \textit{Cross-Algo} contrasts (comparing solutions with different algorithms, $62.8\%$).
Thus, while reasoning is indispensable, it faces limits when navigating intricate code logic or subtle intra class nuances.

Extending the scope to global \textbf{Listwise Ranking} further magnifies this limitation, as Table~\ref{tab:ranking_single_col} reveals a scalability defect where Accuracy@1 drops from the pairwise baseline ($61.3\% \to 31.1\%$) while Spearman Correlation hovers at a notably low level ($\rho \approx 0.23$), indicating that the model \textit{lacks global discrimination capability}, failing to sustain consistency beyond binary interactions.

\paragraph{Finding 3: The ``Implicit World Model'' Leverages Causal Reasoning Beyond Complexity Heuristics and Exhibits Robust Confidence Calibration.}

Tracing the \textit{Complexity Gap} in Figure~\ref{fig:analysis}(f) shows accuracy scales with distinction; crucially, the model's superiority over heuristics in low-gap scenarios proves it detects valid semantic signals rather than simple metrics.
Furthermore, Figure~\ref{fig:analysis}(e) demonstrates \textbf{Calibration}, where confidence correlates strictly with accuracy.
This reliability underpins the Section~\ref{sec:agent} gating mechanism, ensuring agents act with certainty.
\begin{table}
    \centering
    \newcommand{\res}[2]{$#1_{\pm #2}$}
    
    \footnotesize  
    \setlength{\tabcolsep}{3pt} 
    \renewcommand{\arraystretch}{1.25} 
    
    \resizebox{\linewidth}{!}{
        \begin{tabular}{c | c | c c c c}
            \toprule
            \textbf{Size} & \textbf{Corr.} & \multicolumn{4}{c}{\textbf{Accuracy@$k$ (\%)}} \\
            \cmidrule(l){3-6}
            \textbf{($N$)} & \textbf{Spr. $\rho$} & \textbf{$k$=1} & \textbf{$k$=2} & \textbf{$k$=3} & \textbf{$k$=4} \\
            \midrule
            
            \textbf{2} & \res{0.24}{0.01} & \res{61.3}{0.6} & -- & -- & -- \\
            
            \textbf{3} & \res{0.22}{0.00} & \res{43.4}{0.4} & \res{25.5}{0.4} & -- & -- \\
            
            \textbf{4} & \res{0.25}{0.00} & \res{35.0}{1.0} & \res{16.4}{0.7} & \res{10.2}{0.2} & -- \\
            
            \textbf{5} & \res{0.22}{0.00} & \res{31.1}{0.9} & \res{11.2}{0.3} & \res{4.9}{0.1} & \res{3.0}{0.2} \\
            
            \bottomrule
        \end{tabular}
    }
    
    \caption{\textbf{Ranking Performance.} 
    Listwise ranking metrics across varying list sizes $N$. 
    \textbf{Spr.}: Spearman Correlation ($\rho$). 
    \textbf{A@$k$}: Accuracy of the top-$k$ ranking positions (\%).
    ``--'' denotes undefined metrics where $k \ge N$.}
    
    \label{tab:ranking_single_col}
    
\end{table}

\begin{insightbox}
    \textbf{{\textcolor{orange}{\usym{2666}}} Discussion~\ref{sec:agent_motivation}. The Indirect Ceiling of Static Prediction.}
    We interpret 72.2\% as an \textbf{Indirect Epistemic Bound}, constrained by the \textit{Validation-Test Gap} and \textit{Limited Innovative Creativity}.
    Theoretically, since static prediction cannot outperform dynamic verification~\cite{rice1953classesofrecursively}, any gains beyond this saturation point represent the overfitting of distributional noise~\cite{dwork2015reusable}.
    Moreover, this ceiling exposes a deficit in \textit{Generative Innovation}: current LLMs hit a \textbf{Homogeneity Barrier}, producing \textit{functionally isomorphic} solutions that lack context-specific specialization~\cite{anil2024generativeaienhances}.
    Thus, lifting this bound relies on evolving base models to achieve the \textit{Genuine Innovation}.
\end{insightbox}

\subsection{RQ3: Scaling Laws of Data-centric Solution Preference}
\label{subsec:rq3_scaling}

We evaluate the Qwen series across a spectrum from 4B to 1T to determine if predictive capability acts as an emergent scaling property.
Figure~\ref{fig:analysis}(c) details performance on five distinct checkpoints: 4B (\textit{Qwen3-4B-Instruct-2507}), 30B (\textit{Qwen3-Coder-30B-a3b-Instruct}), 235B (\textit{Qwen3-235B-a22b-Instruct-2507}), 480B (\textit{Qwen3-Coder-480B-a35b-Instruct}), and 1T (\textit{Qwen-Max}).

\paragraph{Finding 4: Predictive Accuracy Violates Standard Parameter Scaling Laws.}

Contrary to standard Parameter Scaling Laws, our results (Figure~\ref{fig:analysis}(c)) reveal a \textit{rapid saturation phenomenon}.
Within the Qwen series, performance sees diminishing returns after the initial 30B threshold, creating a statistical plateau that persists even at the 1T scale.
This trajectory implies a distinct ``capacity ceiling,'' suggesting that raw parameter scaling alone is insufficient for further gains in the \textit{Data-centric Solution Preference} task.
In contrast, the distinct superiority of DeepSeek-V3.2 ($61.3\%$) and GPT-5.1 ($58.8\%$) demonstrates that predictive power drives less from raw scale than from \textbf{reasoning-centric architectural paradigms}, implying that future gains will rely on specialized inference incentives rather than simple parameter expansion.


\subsection{RQ4: Comparison with Human Judgment and Validation-Test Gap}
\label{subsec:rq4_cases}

To validate the model's reasoning depth, we conducted a qualitative analysis on the \textit{Google Quest Challenge} from main experiment, which is a multi-label subjective question-answering task.

\paragraph{Finding 5: The Model Outperforms Human Intuition by Rejecting Complexity Bias.}

In the case study of Figure~\ref{fig:case_study_quest}, the model surpassed human judgment by correctly prioritizing a simple LightGBM, whereas humans succumbed to the ``bigger is better'' bias by favoring a complex Deep Neural Network.
It successfully detects small-sample overfitting risks that humans missed, proving that data-grounded reasoning can effectively override superficial human biases.

\paragraph{The Validation-Test Gap.}
We further examine the reliability of execution-based validation metrics ($M_{val}$), derived from internal data splits, as proxies for test performance ($M_{test}$).
As shown in Table~\ref{tab:gap_analysis}, relying solely on $M_{val}$ yields an accuracy of only \textbf{72.2\%}.
This ceiling reveals a substantial \textbf{Validation-Test Gap} stemming from distributional shifts and validation overfitting.
Crucially, implicit reasoning partially mitigates this gap, offering a semantic safeguard that balances efficiency against the risk of metric-driven overfitting.

\begin{table}[h]
    \centering
    \small
    \begin{tabular}{lcc}
    \toprule
    \textbf{Signal Source} & \textbf{Cost} & \textbf{Acc. (\%)} \\
    \midrule
    Random Guess & - & 50.0 \\
    Exec. ($M_{val}$) & $\sim$Hours & 72.2 \\
    \textbf{LLM} & $\sim$Seconds & 61.5 \\
    \bottomrule
    \end{tabular}
    \caption{\textbf{Validation-Test Gap.} Local metrics ($M_{val}$) are noisy proxies for test performance ($M_{test}$), achieving only 72.2\% accuracy due to distribution shifts.}
    \vspace{-15pt}
    \label{tab:gap_analysis}
\end{table}


\section{Agent Integration: \ours}
\label{sec:agent}

Building on the predictive capabilities of the World Model, we propose \textbf{\ours}, a hybrid autonomous ML agent designed to decouple hypothesis exploration from physical execution.

\subsection{Motivation}
\label{sec:agent_motivation}

We aim to break the \textit{Execution Bottleneck} in Section~\ref{sec:motivation}, compressing hours of physical execution into seconds of logical inference, and the \textit{Validation-Test Gap} identified in Section~\ref{subsec:rq4_cases}.
Thus, we propose \textbf{\ours}, which utilizes the ``Implicit World Model'' as a filter to prune the search space before execution for acceleration.

\subsection{Method: The Predict-then-Verify Loop}
\label{sec:method}

We adopt \textbf{AIDE}~\cite{jiang2025aideaidrivenexplorationspace} as our backbone, building directly upon the tree-search architecture described in Section~\ref{sec:agent_paradigm}.

We propose \textbf{\ours}, which re-engineers the Improvement stage into a conservative \textit{Predict-then-Verify} loop (Figure~\ref{fig:main}(e)) to bridge the Implementation Gap~\cite{zhu2025aiscientistsfailstrong}.
The workflow proceeds through three key phases: (1) \textbf{High-Volume Generation}, where $m=10$ candidates are proposed in parallel to expand search width without execution costs; (2) \textbf{Confidence-Gated Pairwise Selection}, which utilizes a confidence gate ($c=0.7$) to ensure high-certainty selection; and (3) \textbf{Verification Execution}, where the Top-$k$ ($k=1$) candidate is physically verified to anchor the solution trajectory in execution feedback (see ablation study in Appendix~\ref{app:ablation_k}).

\begin{figure*}[t]
    \centering
    \includegraphics[width=\linewidth]{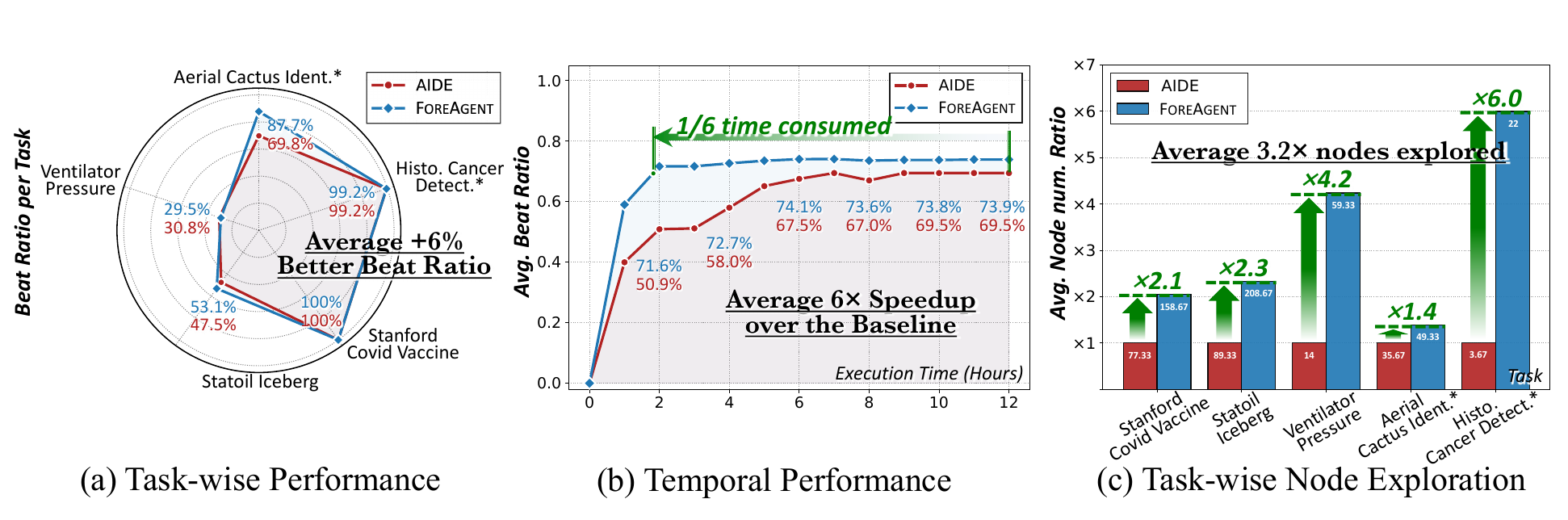} 
    \caption{\textbf{Agent Performance Analysis.}
  (a) \textbf{Task-wise Beat Ratio:} \ours achieves an average +6\% improvement over the AIDE baseline.
  (b) \textbf{Temporal Efficiency:} The agent converges to peak performance using only 1/6 of the execution time, achieving an average $6\times$ speedup.
  (c) \textbf{Search Breadth:} By offloading evaluation to the ``Implicit World Model'', \ours explores $3.2\times$ more nodes on average compared to the baseline, significantly expanding the search space within the same time budget.}
  \label{fig:agent_performance}
\end{figure*}

\begin{table}[h]
\centering
\small
\begin{tabular}{lll}
\toprule
\textbf{Task Name} & \textbf{Domain} & \textbf{Status} \\
\midrule
Stanford Covid Vaccine & Biology & Seen \\
Ventilator Pressure & Physics & Seen \\
Statoil Iceberg & Geoscience & Seen \\
Aerial Cactus Ident.* & Ecology & \textbf{Unseen} \\
Histo. Cancer Detect.* & Medicine & \textbf{Unseen} \\
\bottomrule
\end{tabular}
\vspace{-2pt}
\caption{\textbf{Agent Evaluation Benchmark.} The selection covers diverse AI4Science domains to test the World Model's capability to generalize from seen tasks to unseen scientific problems.}
\label{tab:agent_tasks}
\vspace{-10pt}
\end{table}

\subsection{Experimental Setup}

\paragraph{Tasks and Baselines.} We evaluate \textbf{\ours} on 5 AI4Science tasks from MLE-bench (Table~\ref{tab:agent_tasks}), including two ``Unseen'' tasks.
We benchmark against AIDE under a 12-hour limit; both use DeepSeek-V3.2 for coding, while Implicit World Modeling employs DeepSeek-V3.2-Thinking.

\paragraph{Metric.} To ensure reliability, we conduct three independent runs for each task and report the average \textbf{Beat Ratio}~\cite{ou2025automindadaptiveknowledgeableagent}. This metric quantifies the percentage of human leaderboard contestants outperformed by the agent, representing expert-level competitiveness.

\subsection{Results}
\label{subsec:agent_result}
By substituting costly execution with rapid inference, \ours achieves an average \textbf{$\bm{6\times}$ speedup} (Figure~\ref{fig:agent_performance}(b)), enabling it to explore \textbf{$\bm{3.2\times}$ more nodes} within just 1/6 of the time budget (Figure~\ref{fig:agent_performance}(c)).
This expanded search capability directly translates into performance, driving a \textbf{$\bm{+6\%}$ improvement} in Beat Ratio (Figure~\ref{fig:agent_performance}(a)) and demonstrating robust generalization on unseen tasks.
Furthermore, the World Model acts as a semantic safeguard during intermediate development, significantly boosting the Test Improve Rate by 23\% (see Appendix~\ref{app:fidelity_analysis}).
Although we currently focus on inference, this paradigm naturally extends to training contexts like \textbf{Reward Model}, a promising direction we reserve for future work.

\section{Related Work}
\label{sec:related-work}

\paragraph{LLM Agents in Machine Learning (ML).}
LLM agents are extensively deployed in ML for tasks ranging from pipeline automation~\cite{jiang2025aideaidrivenexplorationspace, qiao2025scalinggeneralistdataanalyticagents, gu2024largelanguagemodelsconstructing} to competitive problem-solving~\cite{luo2025executableknowledgegraphsreplicating, ou2025automindadaptiveknowledgeableagent, chan2025mlebenchevaluatingmachinelearning, liu2025mlmasteraiforaiintegrationexploration}.
However, the computational cost of their generation-execution loops~\cite{yao2023reactsynergizingreasoningacting} remains a bottleneck.
To mitigate this, recent works utilize internal priors to prune redundant steps~\cite{kulibaba2025kompeteaiacceleratedautonomousmultiagent, trirat2025automlagentmultiagentllmframework}, transitioning from brute-force search to reasoned planning.

\paragraph{World Models for Skip-Execution.}
Adapting World Models~\cite{Ding2025, li2025wordworldlargelanguage} to code, recent research predicts execution outcomes to bypass physical runs~\cite{Hora2024predictingtestresultswithoutexecution, faircodegenteam2025cwmopenweightsllmresearch, li2025codeiocondensingreasoningpatterns}.
While prior works focus on logic consistency in reasoning benchmarks~\cite{wei2025equibenchbenchmarkinglargelanguage, gu2024cruxevalbenchmarkcodereasoning, jain2024livecodebenchholisticcontaminationfree}, our approach integrates this predictive capability with Data-Centric Solution Preference~\cite{Shen2024towardsdatacentricrlhf, Just2024datacentrichuman}.
By anchoring evaluations in explicit dataset rationales rather than heuristics, we ensure reliability in stochastic data domains.
Extended discussion in Appendix~\ref{app:extended_related_work}.

\section{Conclusion}



This work validates the feasibility of compressing physical execution into logical inference. Our analysis reveals LLMs function as calibrated, reasoning-driven critics via semantic verbalization to strictly gate actions and prune search spaces. By decoupling reasoning from runtime, we provide a robust blueprint for bypassing the execution bottleneck in complex machine learning tasks.


\section*{Limitations}
\label{sec:limitation}

\paragraph{Corpus Imbalance and Domain Coverage.}
Although our corpus encompasses 18,438 pairs across 26 tasks, the distribution remains inherently skewed.
Mainstream paradigms like Classification and Regression dominate the dataset, whereas niche scientific tasks (e.g., Audio Classification, Tabular Grading) are represented by significantly smaller sample sizes.
Consequently, while the model demonstrates strong general capabilities, its reliability in extremely low-resource or highly specialized scientific domains may vary, and the current evaluation may not fully reflect the challenges of these long-tail scenarios.
Furthermore, the Verified Data Report currently relies on metadata for unstructured domains like CV and NLP, leaving the integration of multimodal data analysis agents for deeper semantic profiling to future work.

\paragraph{Agent Framework Implementation.}
To validate the model's utility, we prioritized stability, instantiating \ours with a conservative \textit{Predict-then-Verify} loop.
This design alternates strictly between singular prediction and execution, barely scratching the surface of potential inference-time strategies.
Specifically, we have not exhaustively explored advanced architectural variants or hyperparameter configurations within this paradigm, implying that the current implementation has not yet been pushed to its optimal limit.
Therefore, the reported performance likely represents a lower bound of the framework's capability.
Beyond this specific instantiation, we identify the framework's broader potential as a scalable \textit{Reward Model}.
By providing dense, execution-free feedback, it paves the way for accelerating Reinforcement Learning rollouts and serves as a plug-and-play optimization module adaptable to diverse agent frameworks.

\bibliography{custom}

\clearpage

\appendix

\section*{Appendix Table of Contents}
\begin{itemize}
    \item \textbf{Appendix \ref{app:extended_related_work}: Extended Related Work}
    \item \textbf{Appendix \ref{apendix:corpus_details}: Corpus Details}
    \begin{itemize}
        \item \ref{sec:appendix_task_detail} Task Metadata and Scale
        \item \ref{sec:appendix_algo_details} Algorithm and Architecture Distribution
        \item \ref{sec:appendix_agent_benchmark} Agent Evaluation Benchmark
        \item \ref{sec:trajectory_sampling} Trajectory Sampling and Intermediate States
    \end{itemize}
    \item \textbf{Appendix \ref{appendix:detailed_exp_result}: Detailed Experiment Result}
    \begin{itemize}
        \item \ref{app:fine_grained_performance} Fine-grained Performance on Prediction Corpus
        \item \ref{app:trajectory_variance} Analysis of Pair Source and Trajectory Variance
        \item \ref{app:ai4s_bench} Detailed Performance Metrics of \ours on AI4Science Benchmarks
        \item \ref{app:search_efficiency} Search Efficiency Analysis of \ours
        \item \ref{app:fidelity_analysis} Decision Fidelity and Reliability in Local Iterations
        \item \ref{app:ablation_k} Ablation Study on Top $k$ Selection
        \item \ref{app:license} Licensing and Artifact Usage
        \item \ref{app:compute} Computational Infrastructure and Budget
        \item \ref{app:packages} Software Dependencies and Metric Implementation
    \end{itemize}
    \item \textbf{Appendix \ref{appendix:case_studies}: Detailed Qualitative Analysis}
    \begin{itemize}
        \item \ref{app:case_bias} Case I: Overcoming Complexity Bias
        \item \ref{app:case_domain} Case II: Domain Fit over Architectural Sophistication
        \item \ref{app:case_report} Case III: Sample of the Verbal Data Report
        \item \ref{app:case_instruction} Case IV: Sample of the Task Instruction ($I$)
    \end{itemize}
    \item \textbf{Appendix \ref{appendix:prompt_templates}: Prompt Templates}
\end{itemize}
\vspace{1em}

\section{Extended Related Work}
\label{app:extended_related_work}

This section expands upon the brief literature review in Section~\ref{sec:related-work}, providing a detailed taxonomy of LLM-based autonomous agents and the theoretical underpinnings of world models in the code domain.

\paragraph{LLM-based Agents for Scientific Discovery}
LLMs with strong reasoning capabilities~\cite{qiao-etal-2023-reasoning} are increasingly serving as core controllers for autonomous agents in scientific discovery~\cite{gu2024largelanguagemodelsconstructing,chen2025ai4researchsurveyartificialintelligence}, extending to specialized machine research domains~\cite{toledo2025airesearchagentsmachine, Wen-lin2025}.
Beyond the digital realm, agents are transforming laboratory research~\cite{liu2025alphagomomentmodelarchitecture, li2025mlrcopilotautonomousmachinelearning, Yuxuan2025, schmidgall2025agentlaboratoryusingllm} and complex data analytics~\cite{sun2025agenticdataagenticdataanalytics, zhang2025deepanalyzeagenticlargelanguage}.
Prominent systems now autonomously propose hypotheses~\cite{chai2025scimastergeneralpurposescientificai, yu2025alpharesearchacceleratingnewalgorithm, internagentteam2025internagentagentscientist, novikov2025alphaevolvecodingagentscientific} and conduct closed-loop experiments~\cite{gottweis2025aicoscientist, lange2025shinkaevolveopenendedsampleefficientprogram, yuan2025dolphinmovingclosedloopautoresearch, yu2025researchtownsimulatorhumanresearch, WOS:001446914500004}, highlighting the trend of ``AI Scientists'' operating in open-ended exploration loops.

Narrowing down to the machine learning domain, the ecosystem is highly diversified.
One stream of research focuses on managing the end-to-end workflow, ranging from autonomous frameworks~\cite{nam2025mlestarmachinelearningengineering, yang2025rdagentllmagentframeworkautonomous, qiao2025scalinggeneralistdataanalyticagents, Ze-xi2025} to engineering pipelines~\cite{Fang2025, chi2024selatreesearchenhancedllm, DOI:10.3724/2096-7004.di.2024.0005}.
Another stream, driven by benchmarks like MLE-bench~\cite{chan2025mlebenchevaluatingmachinelearning, huang2024mlagentbenchevaluatinglanguageagents} and broader evaluation suites~\cite{zhang2025mlrcbenchlanguageagentssolve, jing2024dsbench, zhang2024benchmarking, Nathani2025}, focuses on competitive problem-solving through knowledge-guided reasoning~\cite{luo2025executableknowledgegraphsreplicating, ou2025automindadaptiveknowledgeableagent} and evolutionary optimization~\cite{du2025automlgennavigatingfinegrainedoptimization, guo2024dsagentautomateddatascience, li2025fmagent, liu2025mlmasteraiforaiintegrationexploration}.

Additionally, general-purpose platforms and optimization frameworks offer the foundational tooling and multi-agent architectures required for scalable research~\cite{wang2025openhandsopenplatformai, jiang2025aideaidrivenexplorationspace, hong2024datainterpreterllmagent, qiang2025mledojointeractiveenvironmentsempowering, wang2025configurablemultiagentframeworkscalable}.
However, to mitigate the significant computational overhead of the generation-execution-feedback loop inherent in these systems, recent approaches explore utilizing internal priors to estimate feasibility and prune redundant steps, thereby accelerating optimization~\cite{kulibaba2025kompeteaiacceleratedautonomousmultiagent, trirat2025automlagentmultiagentllmframework, Zhang2024, Astorga2024}.

\paragraph{Operational Details of Agent Baselines}

As introduced in Section~\ref{sec:agent_paradigm}, we take two representative agent frameworks that operate under the \textbf{Generate-Execute-Feedback} paradigm as examples. Here we provide their detailed mechanisms:

\begin{itemize}
    \item \textbf{AIDE:} AIDE~\citep{jiang2025aideaidrivenexplorationspace} is an LLM-based agent that frames machine learning engineering as a code optimization problem.
    It structures the trial-and-error process as a tree search in the solution space, reusing and refining promising code candidates.
    This method effectively trades computational resources for enhanced performance.
    Specifically, AIDE first generates initial code $C_0$ based on instruction $I$.
    The code is executed by training on dataset $D$ to obtain results. Subsequently, AIDE iteratively derives new code $C_1, C_2, \dots, C_t$ based on the feedback.

    \item \textbf{AutoMind:} Building upon the AIDE framework, AutoMind~\citep{ou2025automindadaptiveknowledgeableagent} further integrates a curated expert knowledge base and a self-adaptive coding strategy.
    While retaining the tree search structure, it grounds the agent in domain expertise and dynamically tailors code generation to task complexity.
    This approach aims to reduce invalid attempts by improving the quality of the initial draft and subsequent refinements.
\end{itemize}

\paragraph{World Models and Execution-Free Evaluation}
The concept of World Models originates from model-based reinforcement learning, where agents learn to simulate the environment's transition dynamics to plan actions without expensive trial-and-error~\cite{Ding2025, hafner2024masteringdiversedomainsworld, feng2025webworldmodels, wong2023wordmodelsworldmodels, li2025wordworldlargelanguage}.
Our work adapts this concept to the code generation domain, addressing the ``Execution Bottleneck'' inherent in the agentic loops described above.

Recent research enables models to internalize the execution process, predicting test outcomes~\cite{Hora2024predictingtestresultswithoutexecution, faircodegenteam2025cwmopenweightsllmresearch,wei2025swerladvancingllmreasoning} or assessing logic consistency directly~\cite{li2025codeiocondensingreasoningpatterns, Changshu2024codemind}.
This capability is rigorously evaluated on reasoning-centric benchmarks~\cite{wei2025equibenchbenchmarkinglargelanguage, gu2024cruxevalbenchmarkcodereasoning, jain2024livecodebenchholisticcontaminationfree}.
Unlike traditional benchmarks that may allow for rote memorization, these tasks require models to transcend statistical pattern matching and develop a deep semantic understanding of algorithmic states and control flows~\cite{chen2025scalingagentlearningexperience, sun2025zerosearchincentivizesearchcapability, akhauri2025regressionlanguagemodelscode, Chen2025reval}, serving as the foundational capability for our proposed framework.
Aligning with OpenAI's Level 4 ``Innovators''~\cite{metz2024openai, zhang2025innogymbenchmarkinginnovationpotential}, this empowers agents to drive innovation by leveraging internal world models to proactively prune vast hypothesis spaces, shifting the paradigm from ensuring syntactic correctness to optimizing for semantic success.
This transition resonates with the broader Data-Centric AI movement~\cite{zha2023datacentricartificialintelligencesurvey, cabrera2025machinelearningsystemssurvey}, moving beyond model architecture to focus on the quality of evaluative signals.
Specifically, our framework incorporates rationale-based preference optimization~\cite{Just2024datacentrichuman} and rigorous dataset construction criteria~\cite{Shen2024towardsdatacentricrlhf} to ensure that the ``implicit world model'' is grounded in data-specific realities rather than abstract heuristics~\cite{tschalzev2024datacentricperspectiveevaluatingmachine}.

\section{Corpus Details}
\label{apendix:corpus_details}

To support the reproducibility of our analysis and provide a comprehensive view of the solution space, we provide detailed metadata for the corpus.

\subsection{Task Metadata and Scale}
\label{sec:appendix_task_detail}
Table~\ref{tab:appendix_task_details_part1} outlines the specific characteristics of each of the 26 tasks, including the domain, machine learning paradigm, data size, and the scale of the constructed evaluation set.

\subsection{Algorithm and Architecture Distribution}
\label{sec:appendix_algo_details} 

\begin{figure*}
    \centering
    \includegraphics[width=0.8\textwidth]{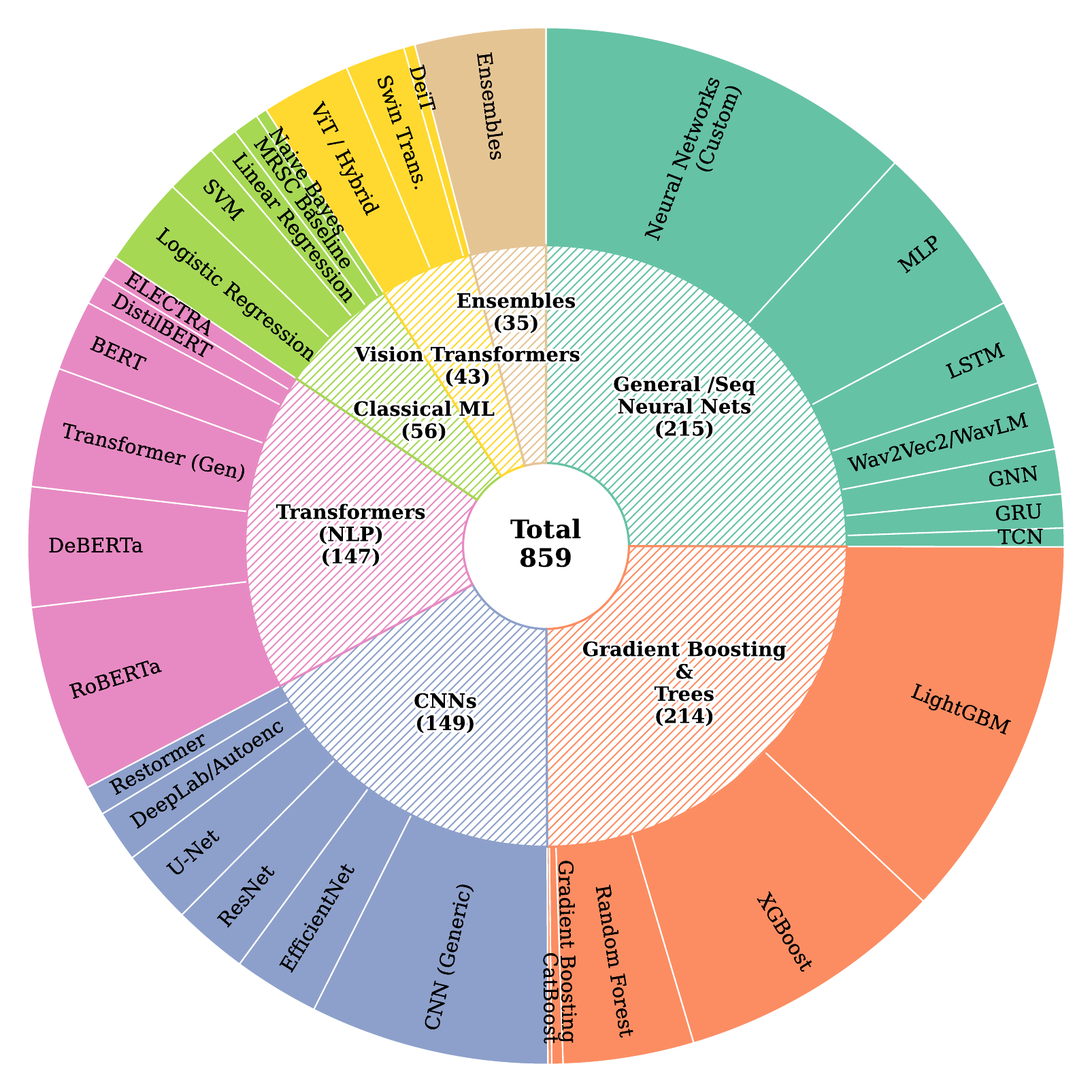} 
    \caption{Hierarchical distribution of the unique solution architectures in our Prediction Corpus. The chart illustrates the balance achieved across major machine learning paradigms: Gradient Boosting\&Trees, General/Sequential NNs, CNNs, and Transformers. The outer ring details specific model instances, demonstrating the high heterogeneity of the solution space.}
    \label{fig:algo_diversity_sunburst}
\end{figure*}

As shown in Figure~\ref{fig:algo_diversity_sunburst} and Table~\ref{tab:appendix_algo_dist_part1}, the solutions range from traditional statistical methods to advanced deep learning architectures, ensuring that our analysis is evaluated against a heterogeneous solution manifold.

\begin{table*}
    \centering
    \footnotesize
    
    \begin{tabular}{
        >{\raggedright\arraybackslash}p{0.18\linewidth} 
        >{\raggedright\arraybackslash}p{0.36\linewidth} 
        >{\raggedright\arraybackslash}p{0.15\linewidth} 
        r r r
    }
    \toprule
    \textbf{Task Name} & \textbf{Task Description} & \textbf{ML Paradigm} & \textbf{Size} & \textbf{Sol} & \textbf{Pair} \\
    \midrule

    \multicolumn{6}{l}{\cellcolor{gray!15}\textbf{\textit{Computer Vision Domain}}} \\
    \midrule
    APTOS 2019 Blindness & Detect diabetic retinopathy severity from retinal fundus images. & Img Class. (Multi-class) & 8.1G & 50 & 1,225 \\ \addlinespace
    
    Dog Breed Identification & Identify dog breed from photos (120 categories). & Img Class. (Multi-class) & 369M & 3 & 3 \\ \addlinespace
    
    Leaf Classification & Classify 99 plant species based on leaf shape features. & Img Class. (Multi-class) & 30M & 17 & 136 \\ \addlinespace
    
    MLSP 2013 Birds & Identify bird species from audio spectrograms. & Img Class. (Multi-label) & 634M & 50 & 1,221 \\ \addlinespace
    
    Plant Pathology 2020 & Distinguish healthy vs. diseased apple leaves. & Img Class. (Multi-class) & 387M & 6 & 15 \\ \addlinespace
    
    Statoil Iceberg Classifier & Distinguish icebergs from ships in radar imagery. & Img Class. (Binary) & 205M & 50 & 1,223 \\ \addlinespace
    
    ICML 2013 Whale & Identify individual Right Whales by callosity patterns. & Img Class. (Multi-class) & 377M & 24 & 275 \\ \addlinespace
    
    TGS Salt Identification & Segment salt deposits from seismic images. & Segmentation (Pixel-level) & 59M & 44 & 880 \\ \addlinespace
    
    \multicolumn{6}{l}{\cellcolor{gray!15}\textbf{\textit{Natural Language Processing Domain}}} \\
    \midrule
    Detecting Insults & Detect insulting language in social commentary. & Text Class. (Binary) & 2M & 27 & 350 \\ \addlinespace

    Jigsaw Toxic Comment & Classify comments into 6 toxicity types (toxic, severe, etc.). & Text Class. (Multi-label) & 129M & 5 & 10 \\ \addlinespace

    Spooky Author ID & Identify author (Poe, Shelley, Lovecraft) of excerpts. & Text Class. (Multi-class) & 3.2M & 50 & 1,220 \\ \addlinespace
    
    Random Acts of Pizza & Predict success of free pizza requests on Reddit. & Text Class. (Binary) & 21M & 50 & 1,225 \\ \addlinespace

    US Patent Matching & Determine semantic similarity between patent phrases. & Matching (Class.) & 316M & 50 & 1,223 \\ \addlinespace

    Denoising Dirty Docs & Restore clean text from noisy scanned documents. & Img Restoration (Reg.) & 97M & 45 & 974 \\ \addlinespace
    
    Google QUEST & Predict 30 subjective attributes (e.g., helpfulness) for Q\&A. & Multi-output (Reg.) & 14M & 50 & 1,224 \\ \addlinespace
    
    Tweet Sentiment Extract & Extract substring supporting the sentiment label. & Seq. Labeling (Extract) & 3.3M & 21 & 210 \\ \addlinespace
    
    LMSYS Chatbot Arena & Predict human preference between two LLM responses. & Ranking (Preference) & 176M & 50 & 1,220 \\ \addlinespace
    
    Automated Essay Scoring & Automatically grade student essays on a numeric scale. & Regression (Ordinal) & 35M & 20 & 190 \\ \addlinespace
    
    \bottomrule
    \multicolumn{6}{r}{\textit{Continued on next page}} \\ 
    \end{tabular}

    \caption{Detailed metadata for all 26 tasks in the Prediction Corpus (Part 1 of 2). The table details the problem definition, ML paradigm, data size, and evaluation scale.}
    \label{tab:appendix_task_details_part1}
    
\end{table*}

\begin{table*}[t!]
    \centering
    \footnotesize
    
    \begin{tabular}{
        >{\raggedright\arraybackslash}p{0.18\linewidth} 
        >{\raggedright\arraybackslash}p{0.36\linewidth} 
        >{\raggedright\arraybackslash}p{0.15\linewidth} 
        r r r
    }
    \multicolumn{6}{c}{{\bfseries \tablename\ \thetable{} -- continued from previous page}} \\
    \toprule
    \textbf{Task Name} & \textbf{Task Description} & \textbf{ML Paradigm} & \textbf{Size} & \textbf{Sol} & \textbf{Pair} \\
    \midrule


    
    
    
    

    \multicolumn{6}{l}{\cellcolor{gray!15}\textbf{\textit{Data Science Domain}}} \\
    \midrule
    NYC Taxi Fare & Predict taxi fare from coordinates and time. & Tabular (Regression) & 5.3G & 30 & 429 \\ \addlinespace
    
    PetFinder Pawpularity & Predict popularity score of pet profile photos. & Regression (Hybrid) & 1.0G & 30 & 239 \\ \addlinespace
    
    NOMAD Conductors & Predict formation energy of aluminum-gallium oxides. & Regression (Scientific) & 25M & 3 & 3 \\ \addlinespace
    
    Stanford COVID Vaccine & Predict degradation rates of mRNA vaccine sequences. & Regression (Bio) & 14M & 50 & 1,222 \\ \addlinespace
    
    Tabular Playground & Predict forest cover type from cartographic variables. & Tabular (Multi-class) & 526M & 24 & 275 \\ \addlinespace
    
    Volcanic Eruptions & Predict time to next eruption from seismic sensors. & Time-Series (Reg.) & 15G & 50 & 1,213 \\ \addlinespace
    
    Ventilator Pressure & Predict airway pressure from control inputs. & Time-Series (Reg.) & 291M & 50 & 1,222 \\ \addlinespace
    
    TF Speech Recognition & Identify spoken commands from audio clips. & Audio (Multi-class) & 2G & 46 & 1,011 \\ 
    \bottomrule
    \end{tabular}
    
    \addtocounter{table}{-1}
    
    \caption{Detailed metadata for all 26 tasks (Part 2 of 2). Continued from previous page.}
    \label{tab:appendix_task_details_part2}
    
\end{table*}


\subsection{Agent Evaluation Benchmark}
\label{sec:appendix_agent_benchmark}

We curated a specialized benchmark to test the World Model's capability to generalize from seen tasks to unseen scientific problems. As detailed in Table~\ref{tab:agent_benchmark}, this selection covers diverse AI4Science domains including Biology, Physics, Geoscience, Ecology, and Medicine. Note that tasks marked with ``*'' (Aerial Cactus and Histo. Cancer Detect) are \textit{unseen} tasks, meaning they were not used in the main experiments and serve as out-of-distribution evaluations.

\begin{table*}[t!]
    \centering
    \footnotesize
    
    \begin{tabular}{
        >{\raggedright\arraybackslash\bfseries}p{0.23\linewidth} 
        >{\raggedright\arraybackslash}p{0.72\linewidth}
    }
    \toprule
    Task Name & Algorithm Composition (Count) \\
    \midrule

    \multicolumn{2}{l}{\cellcolor{gray!15}\textbf{\textit{Computer Vision Domain}}} \\
    \midrule
    APTOS 2019 Blindness & EfficientNet (10), ResNet (10), Swin Transformer (9), ConvNeXt (9), Vision Transformer (8), DeiT (3), CNN-LSTM (1) \\ \addlinespace
    
    Dog Breed ID & ConvNeXt-Large (2), ResNet18 (1) \\ \addlinespace
    
    Leaf Classification & LightGBM (13), Feedforward NN (2), HybridLeafClassifier (1), XGBoost (1) \\ \addlinespace
    
    MLSP 2013 Birds & Ensemble (12), Dual-Stream Arch (5), Feedforward NN (5), Multi-Modal NN (5), Transformer Enc (5), CNN (5), Random Forest (4), XGBoost (4), Logistic Reg (4), LightGBM (1) \\ \addlinespace
    
    Plant Pathology & EfficientNet (2), Swin Transformer (2), ResNet (1), Vision Transformer (1) \\ \addlinespace
    
    Statoil Iceberg & Inverted Bottleneck (5), Vision Trans. (5), ResNet (5), Feedforward NN (5), XGBoost (5), CNN (5), Hybrid CNN-ViT (4), Swin Trans. (4), ConvNeXt (4), Random Forest (3), LightGBM (3), EfficientNet (1), SVM (1) \\ \addlinespace
    
    ICML Whale Challenge & Wav2Vec2 Feature Extractor (10), CNN (6), XGBoost (4), Gradient Boosting (2), LightGBM (1), Mel Spectrogram (1) \\ \addlinespace
    
    TGS Salt ID & Ensemble Segmentation (12), EfficientNet (10), U-Net (10), DeepLabV3Plus (4), Vision Transformer (4), Single Seg. Model (2), Swin Trans. (1), ConvNeXt (1) \\ \addlinespace
    
    Denoising Dirty Docs & Residual Dense Network (10), U-Net (10), Conv Autoencoder (10), Restormer (8), Hybrid CNN-Transformer (6), Simple CNN (1) \\ 
    
    \bottomrule
    \multicolumn{2}{r}{\textit{Continued on next page}} \\
    \end{tabular}

    \caption{Distribution of algorithms and architectures across the corpus (Part 1 of 2). The table details the algorithm composition for Computer Vision tasks.} 
    \label{tab:appendix_algo_dist_part1}
    
\end{table*}

\begin{table*}[t!]
    \centering
    \footnotesize
    
    \begin{tabular}{
        >{\raggedright\arraybackslash\bfseries}p{0.23\linewidth} 
        >{\raggedright\arraybackslash}p{0.72\linewidth}
    }
    \multicolumn{2}{c}{{\bfseries \tablename\ \thetable{} -- continued from previous page}} \\
    \toprule
    Task Name & Algorithm Composition (Count) \\
    \midrule

    \multicolumn{2}{l}{\cellcolor{gray!15}\textbf{\textit{Natural Language Processing Domain}}} \\
    \midrule
    Detecting Insults & DeBERTa (9), Multi-Task DeBERTa-V3 (6), RoBERTa (4), DistilBERT (3), BERT (3), Logistic Regression (2) \\ \addlinespace
    
    Jigsaw Toxic Comment & RoBERTa (3), DistilBERT (1), DeBERTa (1) \\ \addlinespace
    
    Spooky Author ID & Knowledge Distillation (4), DeBERTa (4), ELECTRA (4), BERT (4), LSTM (4), XGBoost (4), Ensemble (4), SVM (4), Logistic Reg (4), Random Forest (3), LightGBM (3), Naive Bayes (3), MLP (2), Transformer (2), Hierarchical Trans. (1) \\ \addlinespace
    
    Random Acts of Pizza & Neural Network (6), SentenceTransformer (4), RoBERTa (4), Knowledge Distillation (4), Multimodal NN (4), BERT (4), DistilBERT (4), Random Forest (4), XGBoost (4), Logistic Reg (4), LightGBM (4), LMs Text Embeddings (4) \\ \addlinespace
    
    US Patent Matching & Custom NN (5), RoBERTa (5), DeBERTa (5), XGBoost (5), BERT (5), Sentence Trans. (5), Similarity Model (5), Linear Reg (5), LightGBM (5), Stacking Ensemble (2), RandomForest (2), Cross-Attn Hybrid (1) \\ \addlinespace
    
    Google QUEST & BERT (5), Multi-Task NN (5), MultiModal Trans. (5), Graph Attention (3), Hierarchical Attn (3), MLP (3), Cross-Attn (3), Sentence Trans. (3), DeBERTa (3), RoBERTa (3), XGBoost (3), LightGBM (3), Ridge Reg. (3), ELECTRA (2), Random Forest (2), LSTM (1) \\ \addlinespace
    
    Tweet Sentiment & RoBERTa-BiLSTM (10), RoBERTa (10), Model Ensemble (1) \\ \addlinespace
    
    LMSYS Chatbot Arena & RoBERTa (11), XGBoost (8), Logistic Reg. (8), LightGBM (8), MLP Classifier (7), DeBERTa (4), Dual Encoder NN (4) \\ \addlinespace
    
    Automated Essay Score & Hybrid NN (9), MetaModel NN (5), Stacking Ensemble (3), LightGBM (3) \\ \addlinespace
    
    \multicolumn{2}{l}{\cellcolor{gray!15}\textbf{\textit{Data Science Domain}}} \\
    \midrule
    NYC Taxi Fare & LightGBM (10), XGBoost (10), Feedforward NN (7), CatBoost (1), Dual-Branch NN (1), Residual NN (1) \\ \addlinespace
    
    PetFinder Pawpularity & LightGBM (27), Vision Transformer (2), XGBoost (1) \\ \addlinespace
    
    NOMAD Conductors & XGBoost (2), Random Forest (1) \\ \addlinespace
    
    Stanford COVID Vac. & Hybrid Architectures (14), Model Ensemble (9), Transformer/GNN (6), Specialized RNA Models (6), Tree Boosters (6), General Baselines (7), LSTM (2) \\ \addlinespace
    
    Tabular Playground & Multi-Branch NN (11), LightGBM (7), Custom NN (3), TabTransformer (2), Feedforward NN (1) \\ \addlinespace
    
    Volcanic Eruptions & Tree Boosters (19), MLP/Dense Networks (16), Transformer Variants (6), CNN/Hybrid Architectures (6), Model Ensemble (2), TCN (1) \\ \addlinespace

    Ventilator Pressure & RNNs (LSTM/GRU) (17), Hybrid Deep Learning (CNN/TCN/Attn) (13), Tree Boosters (10), Transformers (9), Statistical Baseline (1) \\ \addlinespace

    TF Speech Recognition & Statistical ML (RF/SVM/LR) (21), CNN Architectures (13), Pre-trained Audio Models (Wav2Vec2/WavLM) (8), Transformer (2), MLP (1), Knowledge Distillation (1) \\
    
    \bottomrule
    \end{tabular}

    \addtocounter{table}{-1}
    \caption{Distribution of algorithms and architectures across the corpus (Part 2 of 2). Continued from previous page (NLP and Data Science domains).} 
    \label{tab:appendix_algo_dist_part2}
    
\end{table*}

\begin{table*}
    \centering
    \footnotesize
    
    \begin{tabular}{
        >{\raggedright\arraybackslash\bfseries}p{0.20\linewidth} 
        >{\raggedright\arraybackslash}p{0.38\linewidth} 
        >{\raggedright\arraybackslash}p{0.15\linewidth} 
        l 
        c
    }
    \toprule
    Task Name & Task Description & ML Paradigm & Size & Status \\
    \midrule

    \multicolumn{5}{l}{\cellcolor{gray!15}\textbf{\textit{Seen Tasks (In-Distribution)}}} \\
    \midrule
    Stanford COVID Vaccine & \textit{(Biology)} Predict RNA degradation rates at various locations along RNA sequences to assist in mRNA vaccine stability research. & Regression (Seq) & 14M & \textbf{Seen} \\ \addlinespace

    Ventilator Pressure & \textit{(Physics)} Simulate the pressure of a mechanical ventilator connected to a sedated patient's lung to optimize breathing assistance. & Regression (Time-Series) & 291M & \textbf{Seen} \\ \addlinespace

    Statoil Iceberg & \textit{(Geoscience)} Distinguish between icebergs and ships in satellite radar imagery (SAR) to improve navigation safety. & Classification (Image) & 205M & \textbf{Seen} \\ \addlinespace

    \multicolumn{5}{l}{\cellcolor{gray!15}\textbf{\textit{Unseen Tasks (Out-of-Distribution)}}} \\
    \midrule
    Aerial Cactus Identification* & \textit{(Ecology)} Determine the presence of columnar cacti in high-resolution aerial imagery to track protected species in the desert. & Classification (Image) & 25.4M & \textbf{Unseen} \\ \addlinespace

    Histopathologic Cancer Detection.* & \textit{(Medicine)} Identify metastatic cancer tissue in small image patches taken from larger digital pathology scans. & Classification (Image) & 7.7G & \textbf{Unseen} \\
    
    \bottomrule
    \end{tabular}

    \caption{Agent Evaluation Benchmark. The table details the specific tasks used to evaluate the agent, categorized by their domain and their visibility status (Seen vs. Unseen).}
    \label{tab:agent_benchmark}

\end{table*}

\subsection{Trajectory Sampling and Intermediate States}
\label{sec:trajectory_sampling}

To clarify the composition of our Preference Corpus, it is important to note that the dataset captures the entire lifecycle of an agent's exploration.
While we filter out syntactically invalid code that crashes, our dataset is rich in logically imperfect intermediate states. 

As shown in Figure \ref{fig:trajectory_tree}, we generate pairs not just from the final best solution, but across the entire valid search path. 
This extensively tests the model's implicit modeling capabilities to distinguish and guide improvements among messy and unfinalized intermediate code states, ensuring the agent moves from working but poor to working and good.

\newsavebox{\myverbbox}
\begin{figure*}
\centering
\begin{lrbox}{\myverbbox}
\begin{minipage}{0.85\textwidth}
\footnotesize
\begin{verbatim}
Root
 |-- Bug (Syntax Error) [Filtered: Handled by Interpreter]
 |    \-- Bug (Runtime Error) [Filtered]
 |         \-- Score: 0.380 (Valid, Intermediate "Half-baked") -> INCLUDED
 |              \-- Score: 0.377 (Valid, Improved) -> INCLUDED
 |                   |-- Score: 0.385 (Valid, Regressed) -> INCLUDED
 |                   |-- Score: 0.377 (Valid, Best-so-far) -> INCLUDED
 |                   \-- Score: 0.383 (Valid, Sub-optimal) -> INCLUDED
\end{verbatim}
\end{minipage}
\end{lrbox}
\fbox{\usebox{\myverbbox}}
\caption{A representative trajectory snippet illustrating our sampling strategy. This extensively tests the model's implicit modeling capabilities to distinguish and guide improvements among messy and unfinalized intermediate code states, ensuring the agent moves from working but poor to working and good.}
\label{fig:trajectory_tree}
\end{figure*}

\section{Detailed Experiment Result}
\label{appendix:detailed_exp_result}

In this section, we provide a comprehensive breakdown of the experimental results, supplementing the main paper with granular performance metrics across individual tasks, domains, and agent architectures.

\subsection{Fine-grained Performance on Prediction Corpus}
\label{app:fine_grained_performance}
\begin{figure*}
    \centering
    \includegraphics[width=\linewidth]{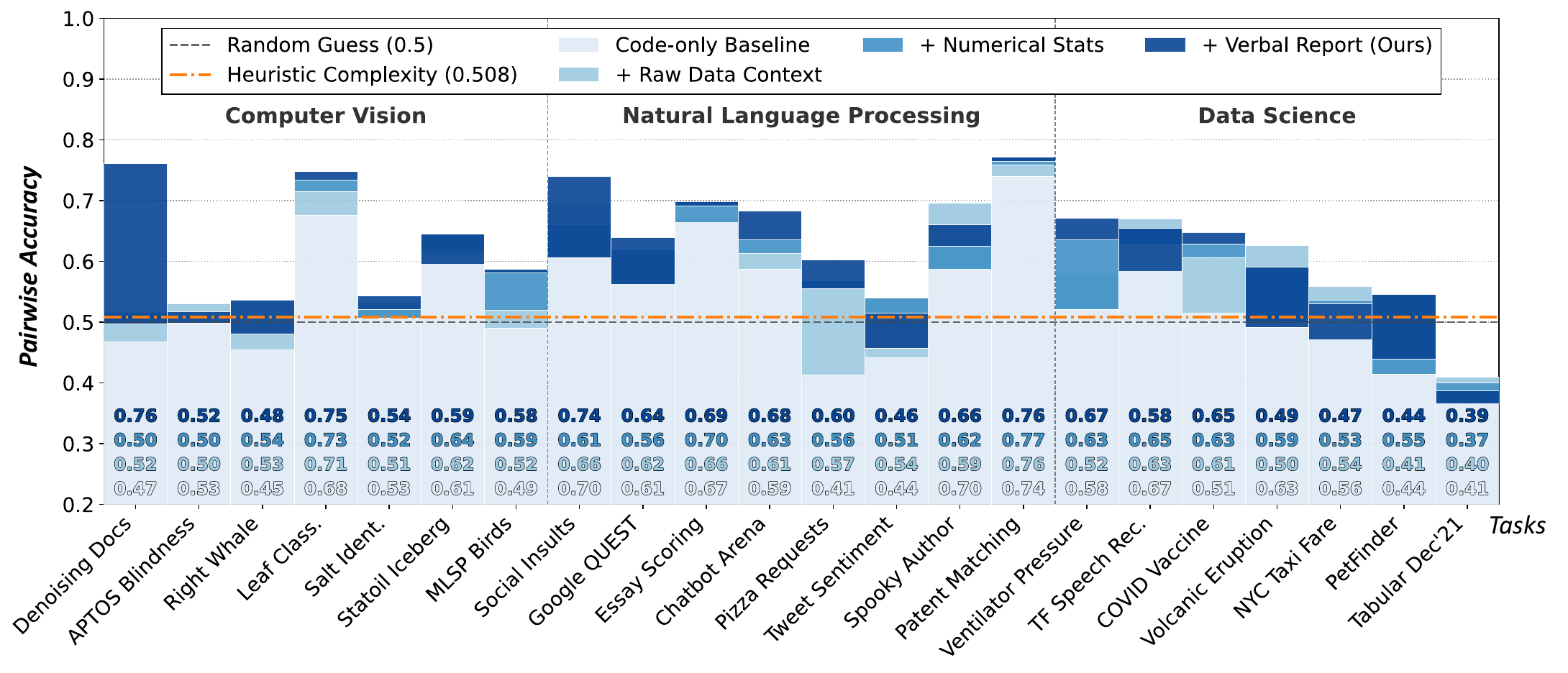}
    
    \caption{\textbf{Domain and Task Sensitivity Analysis.} The stacked bar chart presents the data representation study for each individual task. It visualizes the incremental performance impact of adding Raw Data, Numerical Statistics, and Verbal Reports to the Code-only baseline. The tasks are grouped by their respective domains (CV, NLP, and Data Science) to highlight domain-specific sensitivity.}
    \label{fig:appendix_domain_sensitivity}
\end{figure*}

Table~\ref{tab:appendix_final_v3} presents the task-level performance comparison between DeepSeek-V3.2 and GPT-5.1 across all 26 tasks in the Prediction Corpus.
The results are categorized by task domain (CV, NLP, Data Science) and difficulty level, offering a detailed view of model capabilities.
Furthermore, to provide a deeper understanding of the ``Others'' category mentioned in the main table (Table~\ref{tab:final_compact_matrix_clean}), Table~\ref{tab:appendix_paradigm_breakdown} breaks down performance by specific machine learning paradigms.
This granular analysis reveals distinct performance characteristics in Ranking, Matching, Segmentation, and Extraction tasks, highlighting significant gaps in Matching and Ranking capabilities between the models.
Finally, we investigate the impact of data context in Figure~\ref{fig:appendix_domain_sensitivity}, which presents the data representation sensitivity analysis.
The stacked bar chart reveals the incremental impact of adding Raw Data, Numerical Statistics, and Verbal Reports.
While code-only context serves as a strong baseline, enriching the context with multimodal data yields consistently superior performance, with the magnitude of improvement exhibiting distinct domain-specific patterns.

\subsection{Analysis of Pair Source and Trajectory Variance}
\label{app:trajectory_variance}

To rigorously test whether random sampling introduces easy pairs by comparing an initial half baked script versus a refined solution from the same trajectory, we performed a stratified analysis.
We split our test set into two subsets: Within Trajectory Pairs (both solutions come from the same agent run session) and Cross Trajectory Pairs (solutions come from different agent run sessions or different tasks).

As shown in table~\ref{tab:pair_source}, our experimental results indicate that performance inflation is not observed between two subsets. For DeepSeek V3.2, accuracy on Within Trajectory pairs (60.4\%) is slightly lower than on Cross Trajectory pairs (61.7\%). For GPT 5.1, the performance is statistically identical across both subsets.

A qualitative inspection of the trajectories reveals why the assumption of early half baked code does not hold for our specific agents (AIDE and AutoMind).
Unlike human developers who may write broken snippets in early stages, these autonomous agents are prompted to generate fully executable end to end scripts at every step of the iteration.
Even the first generated solution is typically a complete runnable pipeline.
Later steps represent methodological refinements, such as switching algorithms or adding feature engineering, rather than fixing broken code.

Therefore, distinguishing between an early complete script and a late complete script relies on subtle algorithmic reasoning, not on detecting obvious syntax errors or incompleteness.
The model's predictive capability is robust and reflects genuine reasoning about solution quality rather than the exploitation of trajectory artifacts.

\subsection{Detailed Performance Metrics of \ours on AI4Science Benchmarks}
\label{app:ai4s_bench}

We evaluate the generalization capability of \ours on a subset of 5 challenging AI4Science tasks using the Beat Ratio metric.
Table~\ref{tab:main_results_aide_vs_foreagent} details the specific quantitative results for both the AIDE baseline and \ours.
The comparison explicitly distinguishes between tasks seen during the training phase and unseen out-of-distribution tasks.
The metrics demonstrate that \ours maintains robust performance on seen tasks while achieving superior generalization on unseen problems, such as Aerial Cactus Identification and Histopathologic Cancer Detection, validating the effectiveness of the World Model in bridging the implementation gap.

\subsection{Search Efficiency Analysis of \ours}
\label{app:search_efficiency}

\begin{figure*}
    \centering
    \includegraphics[width=\linewidth]{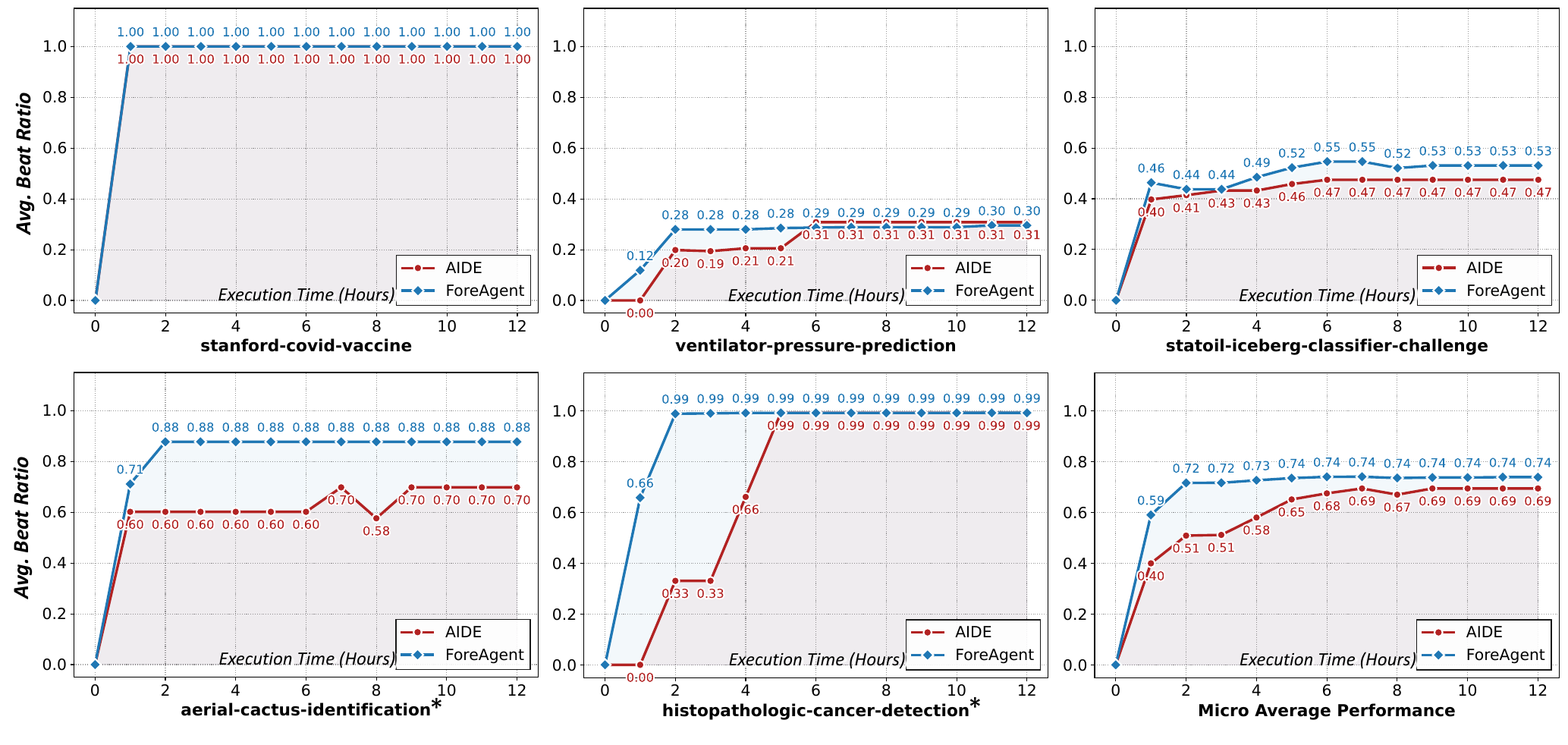}
    \caption{\textbf{Temporal Evolution of Performance.} The curves display the Average Beat Ratio as a function of Execution Time (0--12 hours) for both the AIDE baseline and \textsc{ForeAgent}. The results are broken down by the five individual AI4Science tasks and the overall Micro Average.}
    \label{fig:appendix_agent_perf}
\end{figure*}

\begin{figure*}
    \centering
    \includegraphics[width=\linewidth]{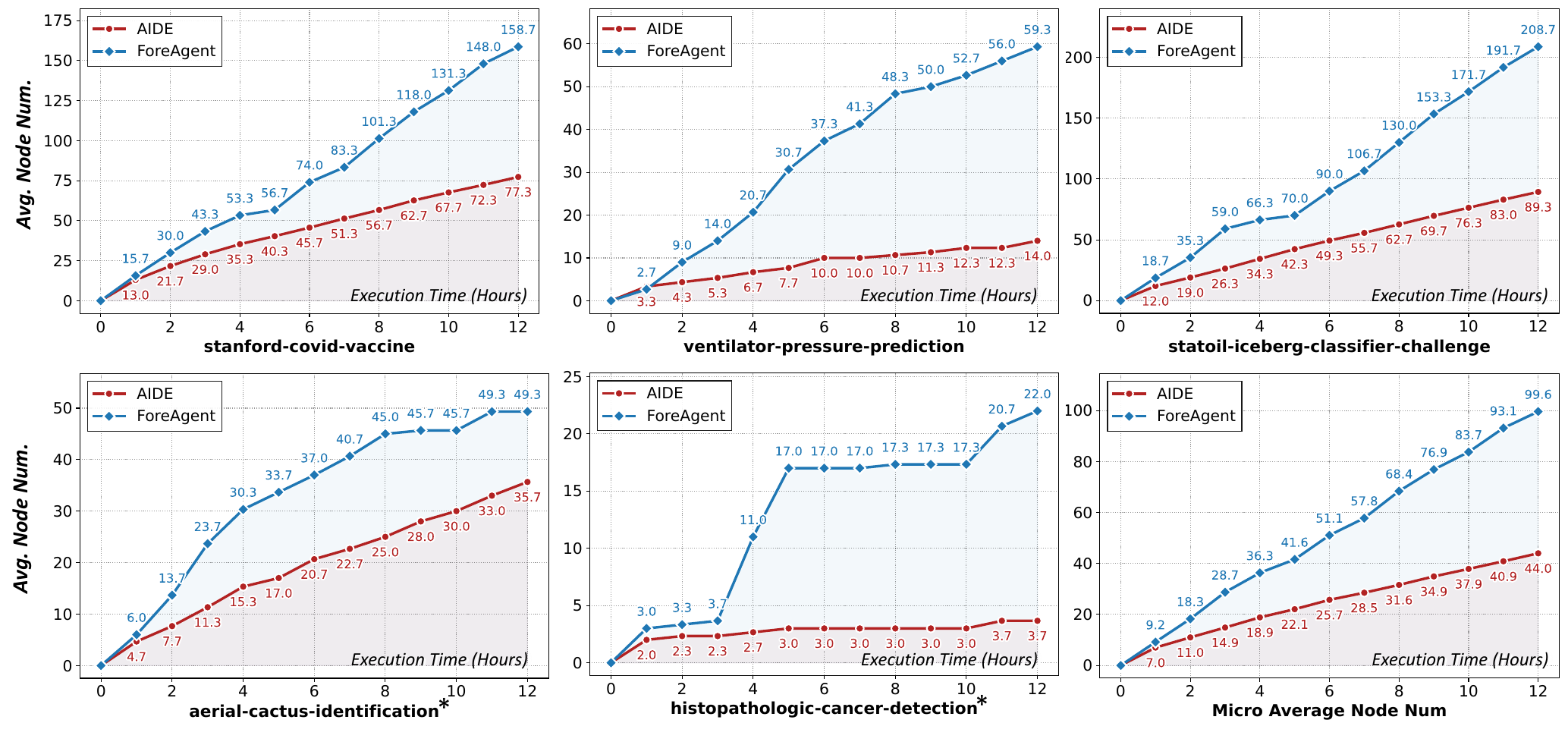}
    \caption{\textbf{Progression of Search Node Exploration.} This figure illustrates the cumulative number of nodes explored (Avg. Node Num.) over the 12-hour duration. It compares the search trajectories of \textsc{ForeAgent} against AIDE across each specific task and the aggregated Micro Average.}
    \label{fig:appendix_agent_node}
\end{figure*}

To elucidate the operational efficiency and robustness of \ours, we analyze its training dynamics.
First, regarding temporal efficiency, Figure~\ref{fig:appendix_agent_perf} plots the Average Beat Ratio over the 12-hour execution window.
The trajectories indicate that \ours converges to optimal solutions significantly faster than the baseline across the majority of tasks.
Complementing this, Figure~\ref{fig:appendix_agent_node} visualizes the search breadth.
It shows that by leveraging the World Model for low-cost evaluation, \ours maintains a higher rate of node exploration, effectively covering a broader search space within the same computational budget.

\subsection{Decision Fidelity and Reliability in Local Iterations}
\label{app:fidelity_analysis}

To investigate whether the framework maintains predictive reliability on noisy intermediate code and prevents the agent from going deeper into incorrect trajectories, we conducted a granular analysis of the decision making process comparing AIDE (execution only) and \ours across 15 runs.
We focused on three key metrics.
Test Improve Rate is the probability that a modification intended to improve the code based on Validation feedback actually yields a better score on the hidden Test set.
Test Non degrade Rate is the probability that the test score of the new candidate is greater than or equal to its parent, indicating the modification did not harm the performance.
Val vs Test Agreement measures how often the Validation signal correctly predicts the Test direction for iterative improvement steps.

As shown in table~\ref{tab:decision_reliability}, the results directly address the impact of intermediate noise.
The execution only baseline (AIDE) struggles with non finalized code, showing a low success rate of 30.39\%.
This indicates that in local iterations, validation scores are extremely noisy.
Traditional agents frequently go deeper into bad trajectories because they are misled by validation overfitting.

\ours significantly improves the Test Improve Rate to 53.49\%.
By introducing the World Model as a filter before execution, the framework successfully prunes many candidate nodes that achieve high validation scores but are logically flawed.
This proves that the World Model does not regress to the baseline when faced with draft code. Instead, it acts as a robust Semantic Safeguard, effectively filtering through the noise of intermediate development and reducing the risk of following incorrect trajectory paths.

Although the Val vs Test Agreement for local improve pairs shows a slight decrease (75.23\%), it remains highly consistent with the global 72\% theoretical ceiling discussed in Section~\ref{subsec:rq4_cases}.
The higher actual success rate demonstrates that \ours is a more reliable navigator for autonomous research and development than execution only feedback.

\subsection{Ablation Study on Top $k$ Selection}
\label{app:ablation_k}

We evaluated the parameter $k$ to balance search breadth and selection stability, comparing our default $k=1$ with $k=2$ and AIDE baseline.

Table \ref{tab:ablation_k} reveals three findings: (1) \textbf{Search Space Expansion}: Setting $k=2$ forces the execution of two candidates per iteration, surging the average node count to 157.87. (2) \textbf{Metric Correlation}: ForeAgent ($k=1$) balances stable Val Test Agreement (75.23\%) with a substantially improved Test Improve Rate (53.49\%). (3) \textbf{Performance Degradation}: Increasing $k$ amplifies exposure to noisy validation signals and increases variance, which destabilizes exploration and lowers the overall score.

In conclusion, $k=1$ provides the optimal balance.
We note this preliminary study only isolates $k$.
Future work will comprehensively explore tuning additional hyperparameter combinations.

\subsection{Licensing and Artifact Usage}
\label{app:license}

We clarify the licensing terms for the key artifacts involved in this study to ensure compliance and reproducibility:

\begin{itemize}
    \item \textbf{Datasets:} All problem statements and datasets are sourced from public Kaggle competitions. They are utilized in strict accordance with their respective competition rules and standard Creative Commons licenses (predominantly CC-BY-SA 4.0).
    \item \textbf{Models:} The backbone language models employed are open-weights models used under their official Apache 2.0 and MIT licenses.
    \item \textbf{Code and Benchmark:} We will release our curated corpus and the accompanying agent framework under the MIT license to facilitate future research.
\end{itemize}

Our usage of these artifacts aligns with their intended purpose of fostering machine learning research.
Furthermore, the derived corpus we will release is strictly intended for non-commercial research evaluation, ensuring compatibility with the original access conditions.

\subsection{Computational Infrastructure and Budget}
\label{app:compute}

\paragraph{Hardware Setup.}
All experiments were conducted on a high-performance local server equipped with an Intel Xeon Gold 6138 CPU (80 logical cores, 2.00GHz) and $6\times$ NVIDIA GeForce RTX 3090 GPUs (24GB VRAM each).
To maximize throughput, we orchestrated a parallelized evaluation pipeline with 6 concurrent workers, assigning one dedicated GPU to each task environment. This ensures that physical code executions are isolated and do not suffer from resource contention.

\paragraph{Token Consumption.}
Table~\ref{tab:compute_budget} summarizes the estimated token usage for the primary data construction and ablation phases.
The main benchmark generation (covering 18,438 solution pairs) consumed approximately 78.5 million tokens (Input + Output).
We note that the computational cost for agent baselines (e.g., AIDE) is highly stochastic due to their autonomous error-recovery loops, where a single difficult task may trigger exponential branching and token usage compared to our linear inference approach.

\begin{table*}
    \centering
    \small
    \begin{tabular}{lcccc}
        \toprule
        \textbf{Experiment Phase} & \textbf{Sample Scale} & \textbf{Input Tokens} & \textbf{Output Tokens} & \textbf{Est. Total} \\
        \midrule
        \textbf{Main Benchmark} & Max 50 sols/task & $\approx$ 60.1M & $\approx$ 18.4M & $\approx$ 78.5M \\
        \textit{(Full Construction)} & (18,438 pairs) & & & \\
        \midrule
        \textbf{Analysis \& Ablation} & Max 15 sols/task & $\approx$ 7.3M & $\approx$ 2.3M & $\approx$ 9.6M \\
        \textit{(Subset Evaluation)} & & & & \\
        \midrule
        \textbf{Agent Baselines} & Dynamic & \multicolumn{2}{c}{\textit{High Variance (Task-Dependent)}} & - \\
        \textit{(AIDE / AutoMind)} & & & & \\
        \bottomrule
    \end{tabular}
    \caption{\textbf{Computational Budget and Token Consumption.} Statistics are aggregated across all 26 tasks. The agent baselines exhibit high variance due to their autonomous feedback loops, making precise token estimation non-deterministic.}
    \label{tab:compute_budget}
\end{table*}

\subsection{Software Dependencies and Metric Implementation}
\label{app:packages}

To ensure the reproducibility of our evaluation metrics and inference pipelines, we detail the software environment and parameter settings used:

\begin{itemize}
    \item \textbf{Evaluation Metrics:} We utilize the standard implementations provided by Scikit-learn for calculating all performance metrics. Unless explicitly stated otherwise, we strictly adhere to the default parameter settings to maintain consistency with standard leaderboards.
    \item \textbf{Data Processing:} Data manipulation and feature extraction are performed using NumPy.
    \item \textbf{LLM Inference:} We employ the official OpenAI Python Library to conduct inference. This standardizes interactions across different model endpoints. We utilize default sampling parameters to ensure deterministic outputs for the "Predict" phase.
\end{itemize}

\newcolumntype{L}[1]{>{\raggedright\arraybackslash}p{#1}}
\newcommand{\res}[2]{$#1_{\pm #2}$}
\newcommand{\bres}[2]{$\bm{#1}_{\pm \bm{#2}}$}

\begin{table*} 
    \centering
    \setlength{\tabcolsep}{5pt} 
    \footnotesize 
    \renewcommand{\arraystretch}{1.15}
    
    \begin{tabular}{L{4.5cm} c c c c c c}
    \toprule
    \textbf{Task Name} & \textbf{Domain} & \textbf{Diff.} & \textbf{Task} & \textbf{Pairs ($N$)} & \textbf{DeepSeek-V3.2} & \textbf{GPT-5.1} \\
    \midrule
    
    APTOS 2019 Blindness & CV & Easy & CLS & 1225 & \bres{51.8}{0.4} & \res{48.2}{1.2} \\
    Denoising Dirty Docs & CV & Easy & REG & 974 & \bres{76.0}{0.6} & \res{53.8}{1.3} \\
    Insults in Social Comm. & NLP & Easy & CLS & 350 & \bres{74.0}{0.3} & \res{60.9}{2.0} \\
    Dog Breed ID & CV & Easy & CLS & 3 & \bres{77.8}{19.2} & \res{66.7}{0.0} \\
    Google QUEST & NLP & Med & REG & 1224 & \res{63.9}{1.1} & \bres{64.6}{0.9} \\
    Jigsaw Toxic Comment & NLP & Easy & CLS & 10 & \bres{23.3}{5.8} & \res{16.7}{5.8} \\
    Leaf Classification & CV & Easy & CLS & 136 & \bres{74.8}{0.4} & \res{72.3}{2.2} \\
    Automated Essay Scoring & NLP & Med & REG & 190 & \res{69.1}{3.6} & \bres{74.5}{1.0} \\
    LMSYS Chatbot Arena & NLP & Med & RNK & 1220 & \bres{68.3}{0.4} & \res{55.8}{0.7} \\
    MLSP 2013 Birds & CV & Easy & CLS & 1221 & \bres{58.1}{1.4} & \res{54.8}{0.5} \\
    NYC Taxi Fare & DS & Easy & REG & 429 & \res{47.1}{1.5} & \bres{52.1}{0.7} \\
    NOMAD2018 Conductors & DS & Easy & REG & 3 & \bres{100.0}{0.0} & \bres{100.0}{0.0} \\
    PetFinder Pawpularity & DS & Med & REG & 239 & \res{43.9}{0.7} & \bres{46.6}{1.2} \\
    Plant Pathology 2020 & CV & Easy & CLS & 15 & \bres{60.0}{11.5} & \res{51.1}{3.8} \\
    Volcanic Eruptions & DS & Hard & REG & 1213 & \res{49.2}{1.2} & \bres{50.5}{0.4} \\
    Random Acts of Pizza & NLP & Easy & CLS & 1225 & \bres{60.2}{0.9} & \res{52.9}{0.6} \\
    Spooky Author ID & NLP & Easy & CLS & 1220 & \res{66.0}{1.0} & \bres{69.2}{1.2} \\
    Stanford COVID Vaccine & DS & Hard & REG & 1222 & \res{64.8}{0.7} & \bres{68.3}{0.3} \\
    Statoil Iceberg & CV & Med & CLS & 1223 & \res{59.5}{1.0} & \bres{62.7}{0.4} \\
    Tabular Playground (Dec) & DS & Easy & CLS & 275 & \res{38.7}{0.4} & \bres{42.7}{1.3} \\
    TF Speech Recognition & DS & Med & CLS & 1011 & \res{58.3}{0.9} & \bres{58.4}{0.4} \\
    TGS Salt ID & CV & Med & SEG & 880 & \res{54.3}{0.7} & \bres{57.9}{0.3} \\
    ICML 2013 Whale & CV & Easy & CLS & 275 & \bres{48.0}{0.4} & \res{47.3}{1.0} \\
    Tweet Sentiment Extr. & NLP & Med & EXT & 210 & \bres{45.7}{3.8} & \res{44.6}{3.1} \\
    US Patent Matching & NLP & Med & MAT & 1223 & \bres{76.4}{0.8} & \res{74.5}{0.2} \\
    Ventilator Pressure & DS & Med & REG & 1222 & \bres{67.0}{0.4} & \res{59.0}{0.4} \\
    \midrule
    \textbf{Overall Average} & \multicolumn{3}{c}{\textit{All 26 Tasks}} & \textbf{18438} & \bres{61.5}{0.2} & \res{58.8}{0.3} \\
    \bottomrule
    \end{tabular}

    \caption{Detailed result of each tasks' performance in the main experiment. Breakdown of Domain, Difficulty (Diff.), and Task Paradigm. $N$ represents the number of pairwise comparison samples. \textit{DS} = Data Science. Values: Mean Accuracy (\%) $\pm$ Stdev. \textbf{Bold}: Best result.}
    \label{tab:appendix_final_v3}
\end{table*}

\begin{table*}
\centering
\footnotesize 
\renewcommand{\arraystretch}{1.15}
\begin{tabular}{lccc}
\toprule
\textbf{Model} & \textbf{Original (Full Set)} & \textbf{Within Trajectory} & \textbf{Cross Trajectory} \\
\midrule
DeepSeek V3.2 & 61.5\% & 60.4\% & 61.7\% \\
GPT 5.1 & 58.8\% & 58.7\% & 58.9\% \\
\bottomrule
\end{tabular}
\caption{Performance Breakdown by Pair Source. Accuracy on Within Trajectory pairs is not inflated.}
\label{tab:pair_source}
\end{table*}

\newcolumntype{L}[1]{>{\raggedright\arraybackslash}p{#1}}

\begin{table*}[t!]
    \centering
    \footnotesize
    \renewcommand{\arraystretch}{1.2} 
    
    \setlength{\tabcolsep}{12pt} 
    
    \begin{tabular}{l c c c}
    \toprule
    \textbf{Task Paradigm} & \textbf{Pairs ($N$)} & \textbf{DeepSeek-V3.2} & \textbf{GPT-5.1} \\
    \midrule
    
    Classification (CLS) & 15,516 & \bres{58.9}{0.3} & \res{57.2}{0.5} \\
    Regression (REG) & 12,685 & \bres{62.1}{0.1} & \res{59.2}{0.3} \\
    Matching (MAT) & 2,356 & \bres{76.6}{0.8} & \res{74.9}{0.3} \\
    Ranking (RNK) & 2,302 & \bres{68.3}{0.4} & \res{55.0}{0.8} \\
    Segmentation (SEG) & 1,639 & \res{54.8}{0.4} & \bres{58.0}{0.2} \\
    Extraction (EXT) & 351 & \bres{46.9}{4.3} & \res{44.1}{3.0} \\
    
    \bottomrule
    \end{tabular}

    \caption{Performance breakdown by specific Task Paradigms. This table expands on the main results by separating the ``Others'' category into Ranking, Matching, Segmentation, and Extraction. Values: Mean Accuracy (\%) $\pm$ Stdev.}
    \label{tab:appendix_paradigm_breakdown}
    
\end{table*}

\begin{table*}
    \centering
    \renewcommand{\arraystretch}{1.2}
    \setlength{\tabcolsep}{8pt} 
    \footnotesize 
    
    \begin{tabular}{l l c c c}
    \toprule
    \textbf{Task Name} & \textbf{Domain} & \textbf{Status} & \textbf{AIDE} (Baseline) & \textbf{ForeAgent} (Ours) \\
    \midrule
    
    \multicolumn{5}{l}{\cellcolor{gray!15}\textbf{\textit{Seen Tasks (In-Distribution)}}} \\
    \midrule
    Stanford COVID Vaccine & Biology & Seen & \bres{1.000}{0.000} & \bres{1.000}{0.000} \\
    Statoil Iceberg Classifier & Geoscience & Seen & \res{0.475}{0.161} & \bres{0.531}{0.134} \\
    Ventilator Pressure Prediction & Physics & Seen & \bres{0.308}{0.041} & \res{0.295}{0.056} \\
    
    \multicolumn{5}{l}{\cellcolor{gray!15}\textbf{\textit{Unseen Tasks (Out-of-Distribution)}}} \\
    \midrule
    Aerial Cactus Identification* & Ecology & Unseen & \res{0.698}{0.157} & \bres{0.877}{0.000} \\
    Histopathologic Cancer Detection* & Medicine & Unseen & \res{0.992}{0.000} & \bres{0.992}{0.001} \\
    
    \midrule
    \textbf{Average Beat Ratio} & \multicolumn{2}{c}{\textit{Across 5 AI4Science Tasks}} & \res{0.695}{0.298} & \bres{0.739}{0.295} \\
    \bottomrule
    \end{tabular}

    \caption{Main results on the MLE-bench AI4Science subset. We report the \textbf{Beat Ratio} (percentage of human contestants outperformed) averaged over 3 independent runs. The ``*'' denotes tasks outside the main evaluation distribution. \textbf{Bold} indicates the best performance.}
    \label{tab:main_results_aide_vs_foreagent}
\end{table*}

\begin{table*}
\centering
\footnotesize
\renewcommand{\arraystretch}{1.2}
\begin{tabular}{lcc}
\toprule
\textbf{Metric} & \textbf{AIDE (Execution only)} & \textbf{\ours (Ours)} \\
\midrule
Test Improve Rate & 30.39\% $\pm$ 11.93\% & \textbf{53.49\% $\pm$ 14.70\%} \\
Test Non degrade Rate & 30.50\% $\pm$ 11.75\% & \textbf{53.97\% $\pm$ 15.00\%} \\
Val vs Test Agreement (Local) & \textbf{78.87\% $\pm$ 10.19\%} & 75.23\% $\pm$ 13.01\% \\
\bottomrule
\end{tabular}
\caption{Decision Reliability in Local Iterations. We report the mean and standard deviation across runs.}
\label{tab:decision_reliability}
\end{table*}

\begin{table*}
\centering
\footnotesize
\renewcommand{\arraystretch}{1.2}
\begin{tabular}{lcccc}
\toprule
\textbf{Model} & \textbf{Mean Score} & \textbf{Node Count} & \textbf{Val Test Agreement} & \textbf{Test Improve Rate} \\
\midrule
AIDE (No WM) & 0.695 $\pm$ 0.298 & 44.00 $\pm$ 33.95 & \textbf{78.87\% $\pm$ 10.19\%} & 30.39\% $\pm$ 11.93\% \\
ForeAgent ($k=1$) & \textbf{0.739 $\pm$ 0.295} & 99.60 $\pm$ 71.49 & 75.23\% $\pm$ 13.01\% & \textbf{53.49\% $\pm$ 14.70\%} \\
ForeAgent ($k=2$) & 0.642 $\pm$ 0.353 & \textbf{157.87 $\pm$ 107.11} & 64.15\% $\pm$ 26.57\% & 35.47\% $\pm$ 23.00\% \\
\bottomrule
\end{tabular}
\caption{Ablation Results on Top $k$ Selection}
\label{tab:ablation_k}
\end{table*}

\section{Detailed Qualitative Analysis}
\label{appendix:case_studies}

\begin{figure*}
    \centering
    
    \begin{CaseStudyFrame}{Case Study: Human Intuition v.s. World Model Inference}{fig:case_study_quest}
        
        \ContextBar{
            \textbf{Task:} Google Quest Challenge (Multi-label Subjective QA Regression). \\
            \textbf{Data Profile:} Small scale ($N_{train} \approx 5.5k$) with 30 \textit{heavily discretized} and \textit{skewed} targets. \\
            \textbf{Data Report Insight:} Strong Question-Group structure (85\% of target variance is explained by Q-means) creates high \textbf{Leakage Risk} in random splits.
        }
        
        \begin{minipage}[t]{0.49\linewidth}
            \begin{SolutionColumn}{myblue!50!black!80}{red!10}{Solution 0: Deep Learning w/ Cross-Attention}{A}
                \small 
                \textbf{Stack:} SentenceTransformer + MultiHead-CrossAttn + AdamW \\
                \textbf{Setup:} Holdout Validation (Random Split), MSE Loss.
                \par\vspace{2pt}
                \textbf{Actual Score:} \textcolor{red!60!black}{\textbf{0.2961} \ding{116}} (Spearman $\rho$)
                
                \vspace{0.2cm} \hrule \vspace{0.2cm}
                
                \HumanView{Neural Network with Cross-Attention explicitly models QA interaction. Pre-trained embeddings capture superior semantics compared to statistical features.}{Strong Favorite \textcolor{red}{\faTimes}}
                \vspace{-2pt}
                
                \WMView{Critique}{Model is underconstrained for small data ($5.5k$). Random split ignores question groups, causing leakage. High overfitting risk.}{Overfitting Risk \textcolor{green!60!black}{\faCheck}}
            \end{SolutionColumn}
        \end{minipage}%
        \hfill
        \begin{minipage}[t]{0.49\linewidth}
            \begin{SolutionColumn}{myblue!50!black!80}{green!10}{Solution 1: LightGBM Ensemble}{A}
                \small 
                \textbf{Stack:} TF-IDF + TruncatedSVD + LightGBM \\
                \textbf{Setup:} 5-Fold Cross Validation + Weighted Ensemble Optimization.
                \par\vspace{2pt}
                \textbf{Actual Score:} \textcolor{green!40!black}{\textbf{0.3145} \ding{115}} (Spearman $\rho$)
                
                \vspace{0.2cm} \hrule \vspace{0.2cm}
                
                \HumanView{TF-IDF is outdated. LightGBM is typically for tabular data, not text. The method is too simple, likely to underfit complex modern natural language processing tasks.}{Weak Baseline \textcolor{red}{\faTimes}}
                \vspace{-2pt}
                
                \WMView{Insight}{Robust choice. 5-Fold CV provides honest estimates. SVD captures global label families. Sample-efficient given data scarcity.}{Optimal Fit \textcolor{green!60!black}{\faCheck}}
            \end{SolutionColumn}
        \end{minipage}
        
        \OutcomeFooter{Solution 1 outperformed Solution 0. The World Model correctly prioritized \textbf{Validation Rigor} and \textbf{Sample Efficiency} over Architectural Sophistication.}
        
    \end{CaseStudyFrame}
    
    \caption{\textbf{Case Study: Human Intuition vs. World Model Inference.} This example illustrates a \textbf{hidden logical conflict} where architectural sophistication (favored by humans) clashes with data constraints. By leveraging the generated \textbf{Data Report}, the World Model detects a critical mismatch between the small dataset size ($N \approx 5.5k$) and the complex neural network (Solution 0). It correctly prioritizes the robust LightGBM ensemble (Solution 1), demonstrating the ability to weigh \textit{Data-Model Fit} over pure algorithmic complexity.}

    \label{fig:case_study_quest}
\end{figure*}

\begin{figure*}[b]
    \centering
    
    \begin{CaseStudyFrame}{Case Study: Prioritizing Domain Fit over SOTA Architecture}{fig:case_study_tgs}
        
        \ContextBar{
            \textbf{Task:} TGS Salt Identification (Pixel-level Segmentation). \\
            \textbf{Data Profile:} Small scale ($101 \times 101$) grayscale seismic images with weak intensity contrast and high variability in mask coverage. \\
            \textbf{Data Report Insight:} High sensitivity to spatial artifacts; resizing introduces interpolation noise that destroys local salt boundary details. Depth data is unavailable (NaN).
        }
        
        \begin{minipage}[t]{0.49\linewidth}
            \begin{SolutionColumn}{myblue!50!black!80}{red!10}{Solution 0: Vision Transformer}{A}
                \small
                \textbf{Stack:} ViT-B/16 (Pre-trained on ImageNet) \\
                \textbf{Setup:} Resize to $224 \times 224$, BCE + Dice Loss, attempts to use depth channel.
                \par\vspace{2pt}
                \textbf{Actual Score:} \textcolor{red!60!black}{\textbf{0.1301} \ding{116}} (mAP)
                
                \vspace{0.2cm} \hrule \vspace{0.2cm}
                
                \HumanView{ViT is a top-tier SOTA model offering global context modeling. Pre-training on ImageNet guarantees strong feature extraction capabilities.}{Strong Favorite \textcolor{red}{\faTimes}}
                \vspace{-2pt}
                
                \WMView{Critique}{Resizing $101 \times 101$ inputs to $224 \times 224$ risks losing fine details. Domain gap between ImageNet (RGB) and seismic data (grayscale) is severe. Incorrectly relies on missing depth data.}{Resolution Mismatch \textcolor{green!60!black}{\faCheck}}
            \end{SolutionColumn}
        \end{minipage}%
        \hfill
        \begin{minipage}[t]{0.49\linewidth}
            \begin{SolutionColumn}{myblue!50!black!80}{green!10}{Solution 1: Standard U-Net}{A}
                \small
                \textbf{Stack:} Custom U-Net with Hybrid Attention \\
                \textbf{Setup:} Native $101 \times 101$ resolution, BCE Loss, Test Time Augmentation (TTA).
                \par\vspace{2pt}
                \textbf{Actual Score:} \textcolor{green!40!black}{\textbf{0.6396} \ding{115}} (mAP)
                
                \vspace{0.2cm} \hrule \vspace{0.2cm}
                
                \HumanView{Standard U-Net is an older, basic baseline. It lacks the global receptive field and advanced attention mechanisms of modern transformers.}{Weak Baseline \textcolor{red}{\faTimes}}
                \vspace{-2pt}
                
                \WMView{Insight}{Preserves native spatial resolution without interpolation noise. Attention mechanisms efficiently capture multi-scale features and spatial dependencies crucial for weak-contrast segmentation.}{Optimal Fit \textcolor{green!60!black}{\faCheck}}
            \end{SolutionColumn}
        \end{minipage}
        
        \OutcomeFooter{Solution 1 outperformed Solution 0. The World Model was not swayed by the ``SOTA'' status, correctly prioritizing \textbf{Native Spatial Resolution} and \textbf{Domain Fit} over superficial architectural complexity.}
        
    \end{CaseStudyFrame}
    
    \caption{\textbf{Case Study: Domain Fit vs. Architectural Sophistication.} This example highlights a ``square peg in a round hole'' scenario. While human intuition might favor the SOTA Vision Transformer (Solution 0), the World Model detects that forcing small $101 \times 101$ seismic images into ViT's required resolution introduces destructive interpolation noise. It rejects the complex model in favor of a U-Net (Solution 1) that preserves native spatial details and avoids relying on missing depth data, demonstrating skepticism towards superficially sophisticated but methodologically flawed approaches.}
    \label{fig:case_study_tgs}
\end{figure*}

\begin{figure*}
    \centering

\begin{ReportFrame}{Case Study: Verbal Data Report ($D_{rep}$) Sample for Task ``US Patent Matching''}{fig:long_report_content_1}

\#\# Data Overview

Train: 32,825 pairs; Test: 3,648 pairs. Columns: id, anchor... score (train only; in \{0.0... 1.0\}). No missing values...

106 unique 4-character CPC contexts; coverage across major sections... broad and similar in train/test.

Anchors: 733 unique; heavily reused (mean \textasciitilde 45 pairs per anchor...). Targets: 26,850 unique...

Test anchors: 100\% seen... Test targets: \textasciitilde 29\% seen... Test OOV rate \textasciitilde 12.5\%.
\newline

Why this structure matters:

The evaluation setting is anchor-centric... rewards learning anchor-specific decision boundaries and context-conditioned mappings...
\newline

\#\# Key Statistical Findings

Discrete labels concentrated at 0.25 and 0.5... Implication: strong class imbalance toward moderate similarity...

Correlation with score: char 3-5gram TF-IDF cosine \textasciitilde 0.46... Implication: surface overlap explains much variance but not all...

Phrases are very short (mean \textasciitilde 2 tokens...)... Implication: models will rely on token/subword and character patterns...

Context means vary... Implication: the mapping from lexical similarity to score is context-dependent...

Anchors average \textasciitilde 45 target pairs... Implication: each anchor induces a nontrivial local decision boundary...

Distributions are stable... Implication: generalization hinges on handling unseen targets...
\newline

\#\# Implications for Model Design

(Linking observation → modeling implication → evaluation impact)

Loss functions that ignore ordinality may mis-penalize near misses... Rank-sensitive metrics will reward monotonic mappings...

Architectures emphasizing character/subword patterns... align with dominant signal... Performance differences will emerge in tail cases...

Pairwise encoders that allow rich anchor-target interactions can learn... Approaches that exploit anchor identity will likely score well...

Models benefit from mechanisms that allow interaction between context and similarity features... Per-context performance may vary...

High-capacity sequence models may be underutilized on such short inputs... Efficiency-capacity trade-offs skew toward models effective on short spans...

Representations that degrade gracefully on unseen words... have an advantage... Robust handling of OOV will particularly improve performance...

Simple identity detection captures some easy gains... however, near-identical forms can still have scores <1.0...

Character CNNs... Token-level transformers... Independent encoders... Joint encoders... Denoising pretraining...
\newline

\#\# Summary

The dataset is anchor-centric... Surface-form similarity explains a large portion... Evaluation emphasizes generalization to new targets...

Capture strong character/subword overlap signals... Maintain robustness to unseen target tokens... Avoid over-reliance on identity heuristics... Account for label ordinality...
    
\end{ReportFrame}
    
\captionof{figure}{\textbf{Case Study: Verbal Data Report ($D_{rep}$) Sample for ``US Patent Matching''.} Generated via the \textit{Code-Execution-Verbalization} protocol, this artifact bridges the gap between raw data statistics and semantic reasoning.}
\label{fig:report_patent}

\end{figure*}

\begin{figure*}
    \centering

\begin{ReportFrame}{Case Study: Task Instruction ($I$) for Task ``Denoising Dirty Docs''}{fig:long_report_content_0}

\# Overview

\#\# Description

Optical Character Recognition (OCR) is the process of converting typed or handwritten documents into a digitized format. OCR makes previously static content editable, searchable, and easier to share.

This task focuses on improving OCR performance for degraded scanned documents. Given a dataset of noisy text images, the goal is to develop a model that removes visual noise such as stains, fading, and wrinkles, producing cleaner text images suitable for further processing and digitization.

\#\# Evaluation

Submissions are evaluated using the root mean squared error (RMSE) between the predicted cleaned pixel intensities and the ground truth grayscale pixel intensities.

\#\#\# Submission Format

Each image is represented as a list of pixel values with identifiers in the form 'image\_row\_col' (e.g. '1\_2\_1' for image 1, row 2, column 1). Intensity values range from 0 (black) to 1 (white).

\verb|```|

id,value
1\_1\_1,1
1\_2\_1,1
1\_3\_1,1
...

\verb|```|

\#\# Data

\#\#\# Dataset Description

The dataset contains two sets of images: 'train' and 'test'.  
- The 'train' set includes noisy text images and their corresponding cleaned versions ('train\_cleaned').
- The 'test' set contains only noisy images that need to be denoised.

The noise simulates real-world artifacts commonly seen in scanned documents, such as blur, stains, and faded ink. The task is to build a model that restores the test images to a clean, readable form.

\#\#\# File Description

- There are three directory corresponding to the data description above: 'train', 'train\_cleaned' and 'test'.
- The sample submission is stored in sampleSubmission.csv.
    
\end{ReportFrame}
    
\captionof{figure}{\textbf{Case Study: Task Instruction ($I$) for Task ``Denoising Dirty Docs''.} 
This example illustrates the raw natural language input $I$ as defined in Section~\ref{sec:agent_paradigm}. 
It outlines the problem context, dataset specifications, and evaluation criteria, serving as the foundational prompt that initiates the agent's solution generation process.}
\label{fig:task_instruction_denoising}

\end{figure*}

To validate the model's reasoning depth and provide transparency into our pipeline, we present four qualitative examples.
First, we analyze two reasoning trajectories (Google Quest Challenge and TGS Salt Identification) to illustrate how the model acts as a skeptical critic, prioritizing methodological fit over superficial sophistication and overcoming human bias (Finding 5).
Subsequently, we provide visual samples of two critical system artifacts: the \textit{Verbal Data Report} and the \textit{Task Instruction}, enabling a concrete inspection of the agent's input and context.

\subsection{Case I: Overcoming Complexity Bias (Reasoning Analysis)}
\label{app:case_bias}

To provide a concrete example of Finding 5 (``The World Model Transcends Human Intuition by Prioritizing Data-Grounded Constraints''), we present a detailed analysis in Figure~\ref{fig:case_study_quest}.
This case illustrates a common pitfall where architectural sophistication clashes with fundamental data constraints.

\paragraph{Scenario and Conflict.}
The agent evaluates two distinct solutions for the Google Quest Q\&A task:
\begin{itemize}
    \item \textbf{Solution 0:} A complex Deep Neural Network (DNN) with Cross-Attention.
    \item \textbf{Solution 1:} A robust LightGBM ensemble.
\end{itemize}
Intuitively, human evaluators and models relying solely on code complexity heuristics often exhibit a ``complexity bias'' by favoring the deep learning approach under the assumption that greater architectural depth yields better performance.

\paragraph{World Model Reasoning.}
However, the World Model leverages the generated Data Analysis Report to detect a critical mismatch.
The report highlights that the dataset is relatively small ($N \approx 5.5k$ samples) with skewed targets.
Synthesizing this finding with model design principles, the World Model predicts a high risk of overfitting for the complex DNN.
Consequently, it correctly prioritizes the LightGBM ensemble (Solution 1), determining that the gradient boosting approach offers a superior Data-Model Fit for this specific sample size.

\subsection{Case II: Domain Fit over Architectural Sophistication}
\label{app:case_domain}

To further substantiate the model's robustness against complex-looking but flawed solutions (``Square peg in a round hole''), we present a second trajectory analysis in Figure~\ref{fig:case_study_tgs}.

\paragraph{Scenario and Conflict.}
The agent evaluates two segmentation solutions for the TGS Salt Identification task:
\begin{itemize}
    \item \textbf{Solution 0:} A Vision Transformer (ViT-B/16) pre-trained on ImageNet.
    \item \textbf{Solution 1:} A standard U-Net.
\end{itemize}
Human intuition might favor the ViT due to its top-tier status and global context modeling capabilities.

\paragraph{World Model Reasoning.}
The World Model recognizes that the task requires pixel-perfect segmentation of small ($101 \times 101$) seismic images. It correctly identifies that forcing inputs into the ViT's required $224 \times 224$ resolution introduces interpolation noise and destroys fine-grained salt details. Unswayed by the ``SOTA'' status, the model rejects the ViT in favor of the standard U-Net, citing the necessity of preserving native spatial resolution and avoiding severe domain mismatch.

\subsection{Case III: Sample of the Verbal Data Report}
\label{app:case_report}

Figure~\ref{fig:report_patent} presents a representative sample of the Verbal Data Report ($D_{rep}$) generated for the \textit{US Patent Matching} task.
This artifact visualizes the mechanism described in Section~\ref{sec:input_augmentation}: transforming raw execution logs (e.g., text length statistics, label skew) into semantic narratives.
It serves as the grounding anchor that allows the language model to ``read'' and internalize dataset properties without direct access to the raw files.

\subsection{Case IV: Sample of the Task Instruction ($I$)}
\label{app:case_instruction}

Finally, to visualize the input definition provided in Section~\ref{sec:agent_paradigm}, Figure~\ref{fig:task_instruction_denoising} displays the raw Task Instruction ($I$) for the task \textit{Denoising Dirty Docs}.
This prompt encapsulates the natural language description, specific dataset paths, and optimization goals, acting as the initial state that triggers the agent's autonomous loop.

\section{Prompt Templates}
\label{appendix:prompt_templates}

To ensure reproducibility and transparency, we provide the full prompt templates used in our World Model framework. The workflow consists of four key stages:

\begin{enumerate}
    \item \textbf{Data Analysis Code Generation (Figure~\ref{fig:appendix_da_code_prompt}):} The agent is first instructed to generate a robust Python script for profiling the dataset. This step extracts key statistical meta-features without training a model.
    \item \textbf{Data Analysis Report Generation (Figure~\ref{fig:appendix_da_report_prompt}):} Based on the execution logs from the previous step, the agent summarizes the findings into a structured, causal report. This report serves as a critical context for the reasoning engine.
    \item \textbf{Result Prediction Query (Figure~\ref{fig:appendix_result_predict_query}):} This is the core reasoning prompt where the World Model predicts the relative performance of candidate solutions. It integrates the task description, the generated data analysis report, and the solution code to form a grounded judgment.
    \item \textbf{Complexity Scoring (Figure~\ref{fig:appendix_complexity_prompt}):} An auxiliary prompt used to calculate the complexity heuristic baseline. It evaluates solutions across code engineering, model architecture, and data pipeline dimensions to detect potential bias towards complexity.
\end{enumerate}

The specific prompt templates are illustrated below.

\onecolumn
\begin{promptbox}{Prompt: Data Analysis Code Generation}
\textbf{SYSTEM:} 

You are an expert Data Science Architect specializing in automated dataset profiling and meta-learning. Your goal is to write a robust, error-handling Python script that extracts high-level statistical and structural insights from a dataset without performing full model training.
\newline

\textbf{USER:}

I need you to generate a Python script to analyze a dataset for the following machine learning task.

Context:

Task Description: \textbf{\{task-desc\}}

Data Directory: \textbf{\{data-dir\}}

Requirements for the Python Script:

Data Loading \& Robustness:
The script must determine the correct data type (Tabular, CV, NLP, or Time-Series) based on the Task Description and file extensions in {data-dir}.
Implement strictly robust file loading (e.g., using try-except blocks). If files are too large, perform stratified sampling (load max 10k rows or 1000 images).

Key Metric Extraction (Crucial for World Model Prediction):
Do not just print raw data. Calculate and print meta-features that correlate with model difficulty.

Output Format:
The script must print the analysis results to stdout in a structured, human-readable text format (or JSON structure) that a downstream LLM can easily parse to write a report.
Do not generate plots/images. Only generate text logs/stats.

Constraints:
Use only standard libraries: pandas, numpy, scipy, sklearn, PIL (for images), os, glob.
Do not attempt to train any machine learning models (e.g., do not run Random Forest).

Response: Provide only the executable Python code block.
\end{promptbox}

\captionof{figure}{Prompt used to instruct the LLM for generating data analysis code.} 
\label{fig:appendix_da_code_prompt} 

\vspace{2em}

\begin{promptbox}{Prompt: Data Analysis Report Generation}

You are preparing a structured Data Analysis Report that will be provided to another expert LLM. Your goal is to make the data characteristics and their implications explicit, causal, and model-relevant — so that a model evaluation agent can reason about how the dataset properties interact with model design choices.
\newline

Follow these instructions carefully:

1. Summarize, don’t just restate numbers.

    - Extract key quantitative trends (e.g., mean intensity, noise variability, contrast, etc.).

    - Highlight patterns, anomalies, and dataset biases.

2. Establish causal implications for modeling.

    - For each key observation, explain why it matters for model training, architecture, or generalization.
    
    - Example: “High inter-sample heterogeneity suggests the model should include normalization or data augmentation to handle distribution shift.”

3. Bridge data to model choices.

    - Express potential advantages or risks for different architectures (CNNs, transformers, denoising autoencoders, etc.) given the observed data patterns.
    
    - DONT directly suggest which model / method is better. You only need to analyze the potential advantages or risks.

4. Directly suggesting models will strongly result in bias.

    - DONT directly suggest which model / method is better. You only need to analyze the potential advantages or risks.

5. Maintain a clear structure using the following format:
\begin{verbatim}
## Data Overview
    <summary of dataset structure, splits, file composition>

## Key Statistical Findings
<highlighted numeric findings + what they imply>

## Implications for Model Design
    <how these data patterns affect likely model performance>

## Summary
    <concise conclusion connecting data traits to modeling priorities>
\end{verbatim}

6. Tone and length:

    - Write concisely and analytically (like a scientific data report).
    
    - Do not include raw metrics dumps.
    
    - Focus on interpretability and causal reasoning.
\newline

Your output will serve as the \{<data\_analysis>\} section for a reasoning-based model evaluator.
Ensure every insight has a clear link from data observation $\to$ modeling implication $\to$ evaluation impact.

INPUT CONTEXT:

[Task Name]
\textbf{\{task\}}

[Task Description]
\textbf{\{desc-block\}}

[Raw Data Analysis Extraction]
\textbf{\{analysis-text\}}
\newline

RESPONSE: Produce the final structured report now. Follow the required headings exactly. Avoid recommending specific models; only analyze potential advantages or risks.
\end{promptbox}
\caption{Prompt used to instruct the LLM for generating data analysis report from the code execution result.} 
\label{fig:appendix_da_report_prompt} 

\vspace{2em}

\begin{promptbox}{Prompt: Result Prediction Query}
     
\textbf{SYSTEM:}

You are an ML code and data analysis expert tasked with predicting the relative performance of provided ML solutions without executing any code. Base your judgment on the task description and the shown code snippets only. Never assume external ground-truth. You should include brief reasoning before the final answer. End your answer with a single JSON object that strictly matches the specified response format.
\newline

\textbf{USER:}

Task: 

\textbf{\{task-name\}}

Task description:

\textbf{\{task-desc\}}

Data analysis:

\textbf{\{data-analysis-report\}}
\newline

Important instructions:

- Predict which solution will perform best (or provide a full ranking) WITHOUT running code.

- Use only the task description, data analysis, and code snippets below.

- Treat the task description and data analysis as equally important to the code; analyze them separately, surface their underlying implications, and provide a balanced, trade-off judgment. 

- Connect data analysis to the following code analysis: If data analysis indicates properties , explain how the architecture addresses them , forming a data→why→method choice causal chain.

- Forbid the “complexity-wins” shortcut: Do not claim “deeper/more complex/with attention is better” as the sole reason. If used, justify why it holds under the current data distribution and training details, and provide a counterexample scenario.

- Response format: \{"predicted\_best\_index": <0 or 1>, "confidence": <optional float>\}

- Indices correspond to the order of the listed solutions (0..n-1).

- You should include brief reasoning before the final JSON. End with a single JSON object matching the response format. Do not write anything after the JSON.
\newline

Provided solutions:

Solution 0: path=\textbf{\{code-0-path\}}

\textbf{\{code-snippet-0\}}

Solution 1: path=\textbf{\{code-1-path\}}

\textbf{\{code-snippet-1\}}

\end{promptbox}
\caption{Prompt used to instruct the LLM for predicting the result of the provided materials.} 
\label{fig:appendix_result_predict_query} 

\vspace{2em}

\begin{promptbox}{Prompt: Complexity Scoring Query}
    
\textbf{SYSTEM:}

You are an expert Machine Learning Engineer and Researcher.
Your task is to analyze a Python script for a machine learning task and evaluate its complexity based on three specific dimensions.

You must output a JSON object with three scores (integers from 1 to 10) and a brief reasoning for each.
\newline

The dimensions are:

1. code\_engineering\_score (1-10): Cyclomatic complexity, custom logic, dependence depth, messy custom loops vs clean API calls.

2. model\_arch\_score (1-10): Parameter count, FLOPs, depth of network, novelty of architecture (e.g., Transformer > Simple CNN).

3. data\_pipeline\_score (1-10): Complexity of preprocessing, data augmentation strategies (Mixup, TTA), custom sampling logic.
\newline

Output Format:

\texttt{\{}

\hspace{1em} "code\_engineering\_score": <int>,

\hspace{1em} "model\_arch\_score": <int>,

\hspace{1em} "data\_pipeline\_score": <int>,

\hspace{1em} "reasoning": "<short summary>"

\texttt{\}}
\newline

\textbf{USER:}

Analyze the following Machine Learning code and provide complexity scores.

\textbf{\{code\_snippet\}}

Respond ONLY with the valid JSON.

\end{promptbox}
\caption{Prompt used to instruct the auxiliary LLM for scoring the complexity of code solutions across three dimensions. This heuristic is used as a baseline to evaluate whether the World Model blindly favors complex code.}
\label{fig:appendix_complexity_prompt}



\end{document}